\documentclass[final,3p,times]{elsarticle}
\AtBeginDocument{%
  \providecommand\BibTeX{{%
    Bib\TeX}}}

\usepackage{fancyhdr}
\usepackage{pdflscape}
\usepackage[table]{xcolor}  
\usepackage{colortbl}       
\usepackage{booktabs}       
\definecolor{LightSkyBlue}{RGB}{135,206,250} 
\usepackage{forest}
\usepackage{xcolor, colortbl}
\usepackage{graphicx}
\usepackage{rotating} 
\usepackage{caption} 
\usepackage{afterpage}
\usepackage{url}
\usepackage{xurl}  
\usepackage{hyperref}  

\usepackage{algorithmic}
\usepackage{textcomp}
\usepackage{overpic}
\usepackage{bbding}
\usepackage{mathtools, nccmath} 
\usepackage{multirow}
\usepackage{multicol}
\usepackage{titlesec}

\usepackage{soul,color,xcolor}      
\definecolor{myColor}{rgb}{0,0,0}        
\makeatletter
\newcommand*{\new}{\@ifnextchar\bgroup{\new@}{\color{myColor}}}
\newcommand*{\new@}[1]{{\textcolor{myColor}{#1}}}
\makeatother

\usepackage{titlesec}
\titlespacing*{\section}{0pt}{1.5ex plus 1ex minus .2ex}{1.3ex plus .2ex}
\graphicspath{{./}}

\def\BibTeX{{\rm B\kern-.05em{\sc i\kern-.025em b}\kern-.08em
    T\kern-.1667em\lower.7ex\hbox{E}\kern-.125emX}}

\def\etc{\emph{etc}}

\definecolor{mydarkgreen}{RGB}{0, 128, 0}
\definecolor{red}{RGB}{255, 0, 0}
\definecolor{blue}{RGB}{0, 0, 255}
\definecolor{gray}{RGB}{220, 220, 220}

\usepackage{stfloats}
\usepackage{fontawesome5}
\def\BibTeX{{\rm B\kern-.05em{\sc i\kern-.025em b}\kern-.08em
    T\kern-.1667em\lower.7ex\hbox{E}\kern-.125emX}}

\begin{document}
\title{From Screens to Scenes: A Survey of Embodied AI in Healthcare}

 \author[label1]{Yihao Liu}
 \affiliation[label1]{organization={The Cancer Research Institute},
             addressline={Central South University},
             city={Changsha},
             postcode={410083},
             state={Hunan},
             country={China}}

\author[label2]{Xu Cao}
 \affiliation[label2]{organization={The Department of Computer science},
             addressline={University of Illinois Urbana-Champaign},
             city={Champaign and Urbana},
             postcode={61801},
             state={Illinois},
             country={USA}}
\author[label3]{Tingting Chen}
 \affiliation[label3]{organization={The the School of Medicine},
             addressline={University of Pennsylvania},
             city={Philadelphia},
             postcode={19104},
             state={Pennsylvania},
             country={USA}} 
\author[label4]{Yankai Jiang}
 \affiliation[label4]{organization={Shanghai AI Lab},
             city={Shanghai},
             postcode={200030},
             state={Shanghai},
             country={China}}
\author[label5]{Junjie You}             
\affiliation[label5]{organization={The School of Life Sciences},
             addressline={Central South University},
             city={Changsha},
             postcode={410083},
             state={Hunan},
             country={China}}

\author[label1]{\\Minghua Wu}
\author[label4]{Xiaosong Wang}
\author[label6]{Mengling Feng}
\affiliation[label6]{organization={The Saw Swee Hock School of Public Health},
             addressline={National University of Singapore},
             postcode={117597},
             country={Singapore}}

\author[label7]{Yaochu Jin}
\affiliation[label7]{organization={The School of Engineering},
             addressline={Westlake University},
             city={Hangzhou},
             postcode={310024},
             state={Zhejiang},
             country={China}}

\author[label8]{Jintai Chen\corref{cor1}}
\affiliation[label8]{organization={The Information Hub},
             addressline={Hong Kong University of Science and Technology (Guangzhou)},
             city={Guangzhou},
             postcode={511453},
             state={Guangdong},
             country={China}}
\cortext[cor1]{Corresponding author (\url{jtchen721@gmail.com}).}

\vspace{-17pt}
\begin{abstract} 
Healthcare systems worldwide face persistent challenges in efficiency, accessibility, and personalization. Modern artificial intelligence (AI) has shown promise in addressing these issues through precise predictive modeling; however, its impact remains constrained by limited integration into clinical workflows. Powered by modern AI technologies such as multimodal large language models and world models, Embodied AI (EmAI) represents a transformative frontier, offering enhanced autonomy and the ability to interact with the physical world to address these challenges. As an interdisciplinary and rapidly evolving research domain, ``EmAI in healthcare'' spans diverse fields such as algorithms, robotics, and biomedicine. This complexity underscores the importance of timely reviews and analyses to track advancements, address challenges, and foster cross-disciplinary collaboration.
In this paper, we provide a comprehensive overview of the ``brain” of EmAI for healthcare, wherein we introduce foundational AI algorithms for perception, actuation, planning, and memory, and focus on presenting the healthcare applications spanning clinical interventions, daily care \& companionship, infrastructure support, and biomedical research, \new{covering 35 specialized tasks}. These significant advancements have the potential to enable personalized care, enhance diagnostic accuracy, and optimize treatment outcomes.
Despite its promise, the development of EmAI for healthcare is hindered by critical challenges such as safety concerns, gaps between simulation platforms and real-world applications, the absence of standardized benchmarks, and uneven progress across interdisciplinary domains. 
We discuss the technical barriers and explore ethical considerations, offering a forward-looking perspective on the future of EmAI in healthcare. A hierarchical framework of intelligent levels for EmAI systems is also introduced to guide further development. By providing systematic insights, this work aims to inspire innovation and practical applications, paving the way for a new era of intelligent, patient-centered healthcare.
\end{abstract}


\maketitle

\section{Introduction}
\thispagestyle{empty}
Healthcare services play a fundamental role in human well-being, yet they face persistent challenges, including inequities in access~\cite{baumann2020reframing}, inefficiencies in care delivery~\cite{wait2017towards}, and a growing demand for personalized solutions to address complex medical conditions~\cite{tortorella2020healthcare, hassan2022innovations}. These issues primarily stem from limited and unevenly distributed healthcare resources~\cite{van2014south}, as well as insufficiently advanced treatment methods~\cite{johnson2021precision}, often resulting in delayed, inadequate, or sometimes excessive treatments that exacerbate patients’ conditions~\cite{may2015burden}. Within the confines of current clinical workflows—largely reliant on finite clinical infrastructure, human healthcare professionals, and caregiving staff—these challenges remain difficult to fully overcome. To address these issues, various efforts have been implemented, such as telemedicine services~\cite{tiwari2023utilization,haleem2021telemedicine}, automated triage systems~\cite{napi2019medical, alhaidari2021triage}, AI-assisted healthcare monitoring~\cite{manickam2022artificial}, and medical image analysis~\cite{liu2023deep, li2024artificial}, which have enhanced the precision and efficiency of medical access. However, they still fall short of providing direct support within existing clinical workflows.

Artificial intelligence (AI) technologies, particularly deep learning approaches, are introducing a new workforce into healthcare practice, driving the ongoing transformation of the healthcare landscape~\cite{yu2018artificial, jiang2017artificial, secinaro2021role, rong2020artificial, aung2021promise, reddy2019artificial, chen2020artificial}. These methods learn medical and diagnostic knowledge from extensive healthcare data collected across multiple centers, scenarios, devices, and time points, utilizing electronic health records (EHRs), genomic sequences, health monitoring signals, and medical images to perform advanced clinical predictive modeling~\cite{shouval2014application, van2022critical}. This enables early-stage diagnoses~\cite{shin2020early}, \new{supports} personalized treatment recommendations~\cite{yogeshappa2024ai}, \new{and detects} subtle disease manifestations beyond human \new{detection}~\cite{chen2024congenital}, collectively improving the quality of healthcare services.

However, the translation of modern AI technologies into tangible clinical benefits remains constrained by at least four fundamental challenges:
(I) Insufficient multimodal processing. Current AI systems primarily rely on one or several common modalities such as vision, language, and audio, but often \new{struggle} to process tactile sensations and olfactory cues, which are both more complex and critical in healthcare. The \new{insufficient} integration for these less-explored modalities limits the effectiveness of AI in addressing the multifaceted nature of clinical tasks and patient care.
(II) The separation between development and deployment. Current deep learning frameworks are characterized by a \new{distinct divide} between development and inference phases, which hinders their continuous \new{adaptation} in real-world clinical settings. This rigid separation delays adaptation to dynamic clinical requirements and evolving environments, ultimately restricting the systems' capacity for ongoing self-improvement.
(III) Insufficient human-machine interaction functionalities. Effective interaction with patients and healthcare professionals is critical for enhancing patient experiences and even improving treatment outcomes. While cutting-edge conversational AI systems, such as ChatGPT and GPT-4, exhibit remarkable interaction capabilities, they often fall short in aligning with \new{clinical goals} and extending beyond verbal communication to encompass behavioral interactions. \new{Effective interactions require} advanced reasoning, robust memory retention, and the ability to adapt based on experience. Although recent studies have highlighted the transformative potential of language in therapeutic contexts~\cite{sedlakova2023conversational}, the mechanisms through which an AI system's linguistic and interactive behaviors can positively impact clinical outcomes—particularly in areas like mental health treatment—remain underexplored~\cite{moreno2020mental, meadows2020conversational}.
(IV) The absence of pathways from decision-making to action execution. Without embodiment in robotic or assistive devices, AI systems are unable to directly alleviate the workload of healthcare professionals and caregiving staff. While current deep learning models may provide accurate diagnoses and decision support, they rarely translate these insights into actionable diagnostic or therapeutic interventions. Furthermore, ensuring safety during such interventions and maintaining seamless integration within established clinical workflows remain critical challenges that require urgent resolution.

Embodied AI (EmAI) is emerging as a promising approach to addressing these challenges in healthcare scenarios~\cite{kaiser2021healthcare, holland2021service, fiske2019your, kruse2023would, huang2023intelligent, kyrarini2021survey}. By integrating AI algorithms, especially multimodal large language models (MLLMs) and world models, with innovations from robotics, mechatronics, human-computer interaction, and sensor technologies, EmAI equips AI algorithms with a physical ``body'', enabling direct interaction with the world~\cite{liu2024aligning}. The AI algorithms are responsible for executing perception, action control, decision-making, and memory processing, ensuring the seamless operation of EmAI systems. Several recent breakthroughs in AI algorithms have significantly advanced the development of EmAI. For instance, unsupervised learning has enabled AI to extract foundational knowledge from vast data without human supervision~\cite{celebi2016unsupervised, dike2018unsupervised, berry2019supervised, balakrishnan2018unsupervised, bommasani2021opportunities}; interactive perceptual learning~\cite{bohg2017interactive} has empowered EmAI systems to comprehend causal relationships of objects and assess the interaction possibilities and feasibility of engaging with various objects~\cite{zhou2023enhancing}; cross-modality fusion techniques have been extensively developed to integrate and leverage complementary information from diverse sources~\cite{cui2023deep, qingyun2021cross}; deep reinforcement learning allows AI systems to learn optimal behaviors through feedback from the environment~\cite{arulkumaran2017deep, franccois2018introduction, ibarz2021train, nguyen2019review, zhu2021deep, zhao2020sim}; and advancements in large language models (LLMs)~\cite{thirunavukarasu2023large, zhao2023survey, chang2024survey, wei2022emergent, gallifant2025tripod}, MLLMs~\cite{yin2023survey, zhang2024mm, mesko2023impact, alsaad2024multimodal, miao2025ultrasound, hu2025adaptlink}, vision-language-action (VLA) models~\cite{brohan2023rt, ma2024survey, kim2024openvla, wen2024tinyvla} and world models~\cite{ha2018world,liu2024world, ding2024diffusion} have provided AI systems with enhanced \new{reasoning, action-planning, and communication capabilities}, particularly for tasks like navigation and manipulation~\cite{zhang2024uni, liu2024vision}. Thanks to these achievements, the development and usability of the ``EmAI brain'' have significantly advanced, enabling more sophisticated, adaptive, and context-aware EmAI systems~\footnote{In this review, we primarily focus on the core component of an EmAI system—the AI system, often referred to as the ``brain'' of EmAI. For the sake of clarity, we do not make a strict distinction between the ``brain'' of EmAI and the overall ``EmAI'' system in the following discussion.} capable of functioning in dynamic healthcare environments.

\begin{figure}[t]
    \centering
    \includegraphics[width=\linewidth]{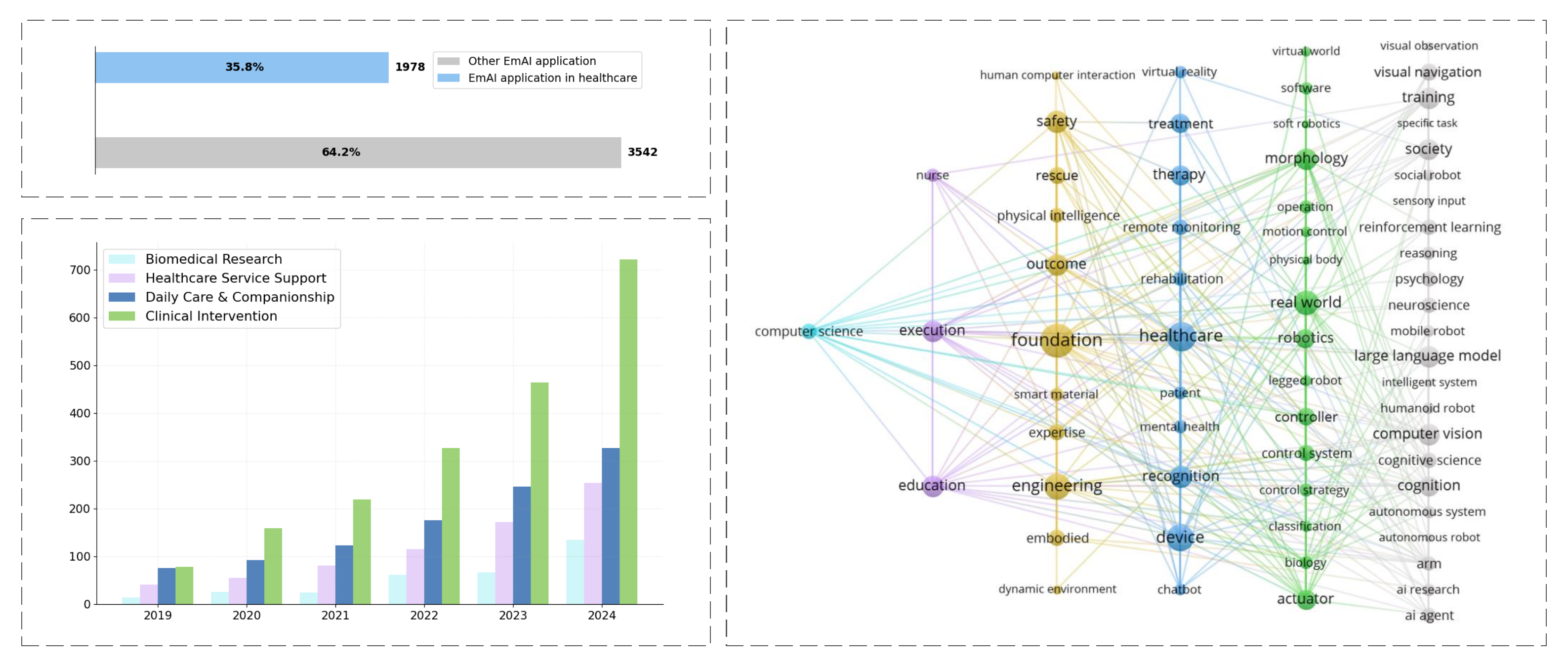}
    \caption{Overview of embodied AI in healthcare research. \textbf{Top left:} The proportion of healthcare-specific EmAI studies among all EmAI publications (as indexed by \textit{Google Scholar}) from 2019 to 2024, reflecting the growing focus on healthcare within this field. \textbf{Bottom left:} Publication trends (2019–2024) for EmAI applications in healthcare. The growing research interest is illustrated across four key areas in healthcare: Biomedical Research, Infrastructure Support, Daily Care \& Companionship, and Clinical Intervention, reflecting the increasing recognition of EmAI's potential to address diverse healthcare challenges. \textbf{Right:} A keyword co-occurrence network was generated using \textit{VOSviewer}~\cite{bukar2023method}, based on \textit{Web of Science} data spanning 2019 to 2024, showcasing core concepts and applications of EmAI in healthcare. Keywords extracted from article titles and abstracts are represented as nodes, with node size indicating frequency and links representing co-occurrence relationships. The network demonstrates a strong trend of interdisciplinary collaboration among fields such as computer science, engineering, and robotics to build holistic EmAI systems for the healthcare applications.}
    \label{fig1}
    \vspace{-7pt}
\end{figure}

Advances in EmAI are driving transformative applications across various fields, with healthcare emerging as a leading domain, which accounts for approximately 35\% of the field's work~\cite{cui2024survey,gao2024empowering}, as illustrated in Figure~\ref{fig1}(a).
Notable examples include surgical robots~\cite{attanasio2021autonomy} and companion robots~\cite{odekerken2020mitigating}, which are becoming increasingly widespread. 
Figure~\ref{fig1}(b) highlights the remarkable growth of EmAI research in key healthcare domains, including biomedical research, infrastructure support, daily care \& companionship, and clinical intervention. Notably, the total number of publications in 2024 is nearly sevenfold that of 2019, with clinical intervention research showing the fastest growth while maintaining a substantial share across these domains.
Such achievements \new{are built on} the integration of insights from multiple disciplines. As depicted in the keyword co-occurrence network (Figure~\ref{fig1}(c)), the dense interconnections across domains \new{reveal} how advancements in one field catalyze progress in others, \new{underscoring} the pivotal role of interdisciplinary collaboration in revolutionizing healthcare. Notable contributions \new{emerge} from breakthroughs in foundation models, large language models, computer vision, cognitive science, sociology, and robotics, collectively shaping the future of EmAI applications in healthcare.
Building on these research achievements, EmAI has been \new{significantly} transforming healthcare by enhancing patient care and operational efficiency. It enables robotic diagnostics~\cite{salcudean2022robot}, precise surgical interventions~\cite{bi2024machine}, and personalized rehabilitation therapies~\cite{lee2024enabling}, \new{which not only streamline} medical workflows \new{but also enhance} health outcomes and reduce recovery times~\cite{wang2023accelerating}. Beyond clinical applications, EmAI provides meaningful companionship~\cite{breuer2023engineers} and emotional support~\cite{ben2020impact}, \new{which is especially beneficial} to vulnerable groups such as children, the elderly, and individuals with disabilities or chronic illnesses, \new{thus helping to ease} the burden on healthcare providers.
In addition, EmAI is \new{transforming} biomedical research by automating experimental processes and analyzing large-scale datasets, \new{which allows} researchers to generate insights and conduct experiments with unprecedented speed. These advancements have accelerated the discovery of medical mechanisms~\cite{Bhinder2021Artificial, Klauschen2023Toward, Shams2024Leveraging}, therapeutic targets~\cite{Pun2022Identification, Bhattamisra2023Artificial}, and disease prevention strategies~\cite{Sak2021Artificial, Istasy2022The}, driving innovation across the biomedical landscape.

Despite significant advancements~\cite{huang2023intelligent, pee2019artificial, fiske2020implications, kokkonen2023beyond, soljacic2024robots, kavidha2021ai, okamura2010medical, terry2019regulating}, the development of EmAI for healthcare remains in its infancy and faces multiple challenges. Current efforts often concentrate on isolated components of EmAI~\cite{wang2023accelerating, kalaivani2023cognitive, weng2020embodiment}, such as developing advanced algorithms~\cite{bachem2023ai, shin2023embodying}, improving workflows~\cite{deitke2022️, xu2024transforming}, or curating data sets~\cite{mlakar2021multilingual, kumar2021study}, without achieving integration into comprehensive systems. To realize the full potential of EmAI, cross-disciplinary collaboration is essential to bridge these fragmented contributions and build cohesive, end-to-end solutions.
Moreover, research has \new{largely} focused on high-profile applications, such as surgical robotics~\cite{salcudean2022robot, hidalgo2023current, jiang2023skeleton, bi2024machine, liu2023robotic, lin2024cfanet, pore2021safe, su2024fully}, while other promising areas, \new{such as} mental health interventions~\cite{fiske2020implications, kokkonen2023beyond}, remain underexplored. This uneven distribution of attention limits the broader impact of EmAI across diverse healthcare needs. Additionally, while companion robots have shown potential, most are reactive rather than proactive~\cite{breuer2023engineers}, \new{which restricts} their ability to anticipate and address patient needs autonomously~\cite{bilalpur2024learning, li2023systematic}. Similarly, biomedical research robots face \new{challenges} in maintaining precision and reliability in the inherently complex and dynamic environments of medical research.

\begin{figure}[t!]
    \centering
    \includegraphics[width=\linewidth]{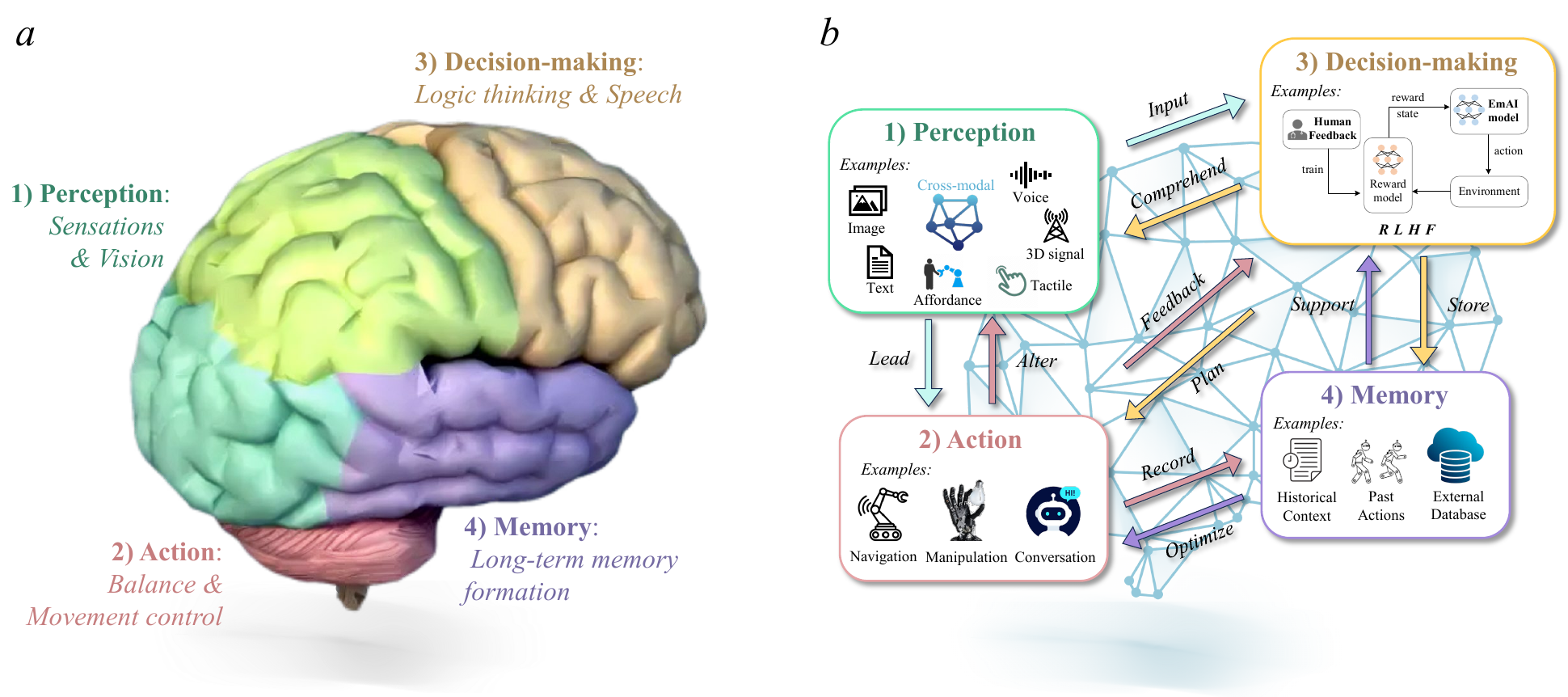}
    \caption{An overview and the function of embodied AI's ``brain". (a) The human brain comprises specialized regions responsible for perception (e.g., parietal and occipital lobe), action control (e.g., cerebellum), decision-making (e.g., frontal lobe), and memory (e.g., hippocampus). (b) Similarly, the ``brain" of embodied AI system is designed to emulate these functions, with interconnected modules for multi-modality perception, decision-making, action control, and memory. These components work synergistically, exchanging feedback and supporting adaptive behavior.}
    \label{fig2}
    \vspace{-7pt}
\end{figure}

Additionally, the development of EmAI for healthcare \new{still faces} significant technical challenges. First, EmAI development is typically carried out on simulation platforms, which often fail to replicate real-world environments \new{accurately}. This discrepancy \new{poses} a major challenge in bridging the gap between simulations and real-world applications. Additionally, as EmAI systems may directly interact with the real world, ensuring safety becomes even more critical, especially in medical tasks~\cite{zhou2024addressing, chaby2022embodied, zhang2025adg}. Second, although EmAI systems rely on large datasets, the acquisition of large, ethically sourced, domain-specific real-world datasets in healthcare \new{is obstructed} by privacy regulations and complex clinical workflows, creating significant barriers to the development of healthcare-specific EmAI. Other challenges, such as ethical considerations~\cite{elendu2023ethical, de2024disability, boada2021ethical, stahl2016ethics} and economic and societal implications~\cite{maibaum2022critique, fosch2021healthcare},\new{must also} be addressed.

Given the promising potential and numerous benefits of EmAI for both patients and healthcare professionals, as well as the existing challenges, a timely summary of these aspects is crucial for advancing the field and fostering interdisciplinary collaboration. In this review, we summarize and discuss recent and emerging applications of EmAI in healthcare, \new{emphasizing} key factors that could significantly impact patient outcomes and healthcare practices. In Section~\ref{section2}, we provide a concise overview of the technologies that underpin the ``EmAI brain'', covering four essential capabilities: perception, actuation, high-level planning, and memory. While our focus is not on \new{exploring} the technical foundations of EmAI (for technical reviews, see \cite{zheng2024survey, zeng2023large, ma2024survey}) or its general applications in robotics (refer to \cite{duan2022survey, pfeifer2004embodied, liu2024aligning}), we present a comprehensive review centered on healthcare applications of modern EmAI, particularly in clinical interventions, daily care \& companionship, infrastructure support, and biomedical research (discussed in Section~\ref{section3}). We also summarize their progress and limitations, categorizing EmAI into five levels of intelligence and illustrating each with examples from various healthcare domains (see Section~\ref{section4}). This framework aims to guide researchers and practitioners in understanding the evolution and stages of EmAI in healthcare. Datasets and benchmarks for diverse healthcare scenarios are summarized in Section~\ref{section5}, while the challenges and opportunities are further outlined in Section~\ref{section6}, with the aim of guiding researchers towards relevant fields, applications, and data foundations for future exploration.

\new{\textbf{Highlights.} This survey is \textbf{the first} to comprehensively integrate technological, application-driven, and ethical perspectives on Embodied AI in healthcare. Specifically, we:
\begin{itemize}
    \item \textit{\textbf{Techniques:}} Provide an in-depth review of foundational algorithms for multimodal perception, actuation, planning, and memory, which form the core of EmAI’s intelligent systems.
    \item \textit{\textbf{Applications:}} Examine how EmAI is revolutionizing healthcare across four critical domains, covering 35 specialized tasks and underscoring its transformative value in diverse scenarios.
    \item \textit{\textbf{Datasets and Benchmarks:}} Summarize and categorize key datasets and benchmarks to facilitate and accelerate research in diverse application fields.
    \item \textit{\textbf{Evaluation Framework:}} Propose a novel framework for assessing the intelligence levels of EmAI systems, guiding future development.
    \item \textit{\textbf{Challenges and Opportunities:}} Discuss technical barriers, safety concerns, and ethical considerations, offering a forward-looking perspective on the future of EmAI in healthcare.
\end{itemize}}

\noindent
\begin{minipage}[b!]{0.62\textwidth}
    \vspace{12pt}
    \section{Basic AI Techniques for Embodied AI}
    \label{section2}
    \hspace{12pt} EmAI is gaining momentum thanks to advancements across multiple fields, particularly \new{recent} breakthroughs in AI. To ultimately replicate human-like behavior in real-world contexts~\cite{zador2023catalyzing}, a comprehensive EmAI ``brain" should encompass multiple modules \new{that} conduct perception, action control, decision-making, and memory. Similar to the human brain, which consists of several specialized but interconnected functional regions (see Figure~\ref{fig2}(a)), these integrated capabilities enable EmAI systems to interact with and adapt to complex real-world environments~\cite{WorldLabs2024SpatialIntelligence, batra2020rearrangement, francis2022core, duan2022survey}, as illustrated in Figure~\ref{fig2}(b).
    Here, we outline key approaches that support these functions, categorized into embodied perception, low-level actuation, high-level planning, and memory processing, along with their breakdowns \new{in detail}, as shown in Figure~\ref{fig_breakdown}. We will also summarize the major achievements \new{in these areas} in this section.
\end{minipage}\hfill
\begin{minipage}[b!]{0.36\textwidth}
    \vspace{20pt}
    \centering
    \includegraphics[width=\textwidth]{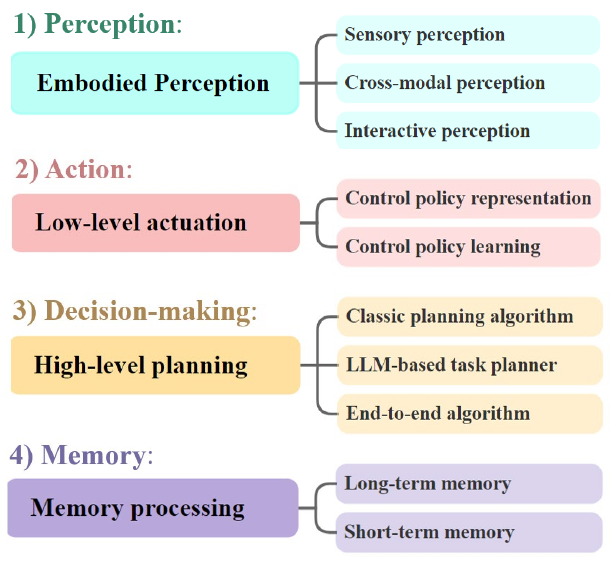}
    \captionsetup{skip=5pt} 
    \captionof{figure}{A detailed breakdown of EmAI core functionalities, with key approaches that support them.}
    \label{fig_breakdown}
\end{minipage}

\subsection{Embodied Perception}
Perception is a core mechanism by which EmAI systems interpret sensory data from their environment. This process involves handling high-dimensional, multimodal, and often noisy inputs from sensors such as cameras, microphones, and tactile devices. This section \new{divides} embodied perception from three key aspects: sensory perception, cross-modal perception, and interactive perception. Sensory perception forms the basis for other system functionalities and integrates directly with existing single-modality foundation models~\cite{wang2023review, pinto2023tuning, yuan2021florence, borsos2023audiolm, yang2023uniaudio, latif2023sparks, liu2022audio, zhao2023survey, chang2024survey, wang2023accelerating}. To achieve a richer understanding of the environment, multimodal AI algorithms~\cite{zhang2024deep, carolan2024review, he2020introduction, wei2021ai} enable \new{the integration of cross-modal information}, aligning with the inherently multimodal nature of the real world. The \new{perception} of multimodal data from different devices allows robots to combine sensory inputs such as vision, touch, and speech to make more informed decisions~\cite{qi2022multimodal}. Interactive perception, which further \new{explores} object affordances, represents a pivotal step in bridging perception and action, \new{making it a significant area for future development in EmAI perception}. We will delve into cross-modal and interactive perception in greater detail in the following sections.

\subsubsection{Cross-modal Perception}
Cross-modal perception integrates information from multiple modalities to achieve a holistic understanding. To efficiently aggregate and align multimodal information, current pre-trained models establish foundational cross-modal representations, enabling downstream multimodal tasks \new{such as} Visual Language Navigation (VLN) and Embodied Visual Question Answering (VQA), \textit{etc}. Recent studies~\cite{zhou2024uncertainty, Li2024Multi-modal, kim2021vilt,stelzner2024recognition,turkoglu2022film,Li2024StitchFusion,jia2021scaling, nan2021joint} predominantly adopt three primary architectural paradigms to achieve effective cross-modal perception:

One prevailing strategy employs \textit{\textbf{separate encoders}}, where each modality is processed independently before fusion. The similarity among cross-modal representations is then computed and optimized to project multimodal information into a shared representation space, often achieved using contrastive loss functions~\cite{liu2021contrastive, liu2023attention, yang2022mcl, tang2024hierarchical}. Taking two two-modality processing as an example, representative dual-encoder models include CLIP~\cite{radford2021learningtransferablevisualmodels}, ViLT~\cite{kim2021vilt}, and ALIGN~\cite{jia2021scaling}.
Additionally, some approaches align multimodal representations with language representations, positioning language as a central anchor to bridge diverse modalities (e.g., video, audio, etc.) and ensure semantic consistency~\cite{zhu2023languagebind}. In most cases, the language encoder is pre-trained on large-scale datasets and remains fixed, while each modality is assigned a dedicated upstream encoder for semantic alignment. This design effectively preserves the model's ability to incorporate new modalities, making it particularly suitable for tasks requiring strong semantic alignment, such as zero-shot learning and cross-modal retrieval. Some approaches~\cite{tsai2024text} leverage pre-trained MLLM to convert all modalities to texts before encoding them, which makes unseen modalities can be efficiently dealt with in inference.
However, due to the relatively shallow level of modality interaction, separate encoder frameworks exhibit limited performance in complex scene understanding tasks~\cite{bao2022vlmo, radford2021learning, wang2021distilled, salemi2023symmetric, jiang2020dualvd}.

Deeper cross-modality interaction is often achieved by employing a \textit{\textbf{shared encoder}} to learn comprehensive cross-modal representations. These shared encoders typically leverage multi-layer Transformers that encode multimodal inputs through representation fusion techniques such as cross-attention mechanisms~\cite{gonccalves2022survey, zheng2022cross, nan2021joint, niu2021review} or feature-wise linear modulation (FiLM)~\cite{stelzner2024recognition, brockschmidt2020gnn, turkoglu2022film}. Representative shared encoder frameworks include ViLBERT~\cite{lu2019vilbert}, LXMERT~\cite{tan2019lxmert}, and UNITER~\cite{chen2020uniter}.
By using deep fusion encoders, this architecture learns more generic cross-modal representations and achieves superior pre-training efficacy for visual reasoning tasks~\cite{zhou2020unified, su2019vl, li2020unimo, li2020oscar} and few-shot tasks~\cite{azeem2024unified, lu2021novel, liu2024few, zhou2024less, wang2024data}. However, it introduces quadratic (for two modalities) or cubic (for three modalities) time complexity, as it requires interaction between all possible modality pairs. This results in significantly slower inference speeds, \new{which limits} its practical applicability. Moreover, bridging the cross-modality gap with a single parameterized model remains challenging, further \new{obstructing} its deployment in real-world scenarios.

Therefore, a combination of the aforementioned approaches, referred to as the \textbf{\textit{combination architecture}}, has also been proposed. Typically, these methods employ separate encoders for modality-specific feature extraction alongside a shared encoder for joint feature learning. Furthermore, \textit{\textbf{techniques aimed at reducing computational overhead}}, such as Mixture-of-Modality-Experts (MoME)~\cite{bao2022vlmo, Li2024Multi-modal, lin2024moma, lin2024moe}, Mixture-of-Prompt-Experts (MoPE)~\cite{Jiang2023Conditional}, effective self-attention~\cite{zhao2020exploring, liu2023sfusion}, and Multi-directional Adapter~\cite{Li2024StitchFusion}, \new{use} sparse routing or modular expert networks to efficiently manage modality-specific information while minimizing computational costs.

\subsubsection{Interactive Perception} 
Interactive perception involves physical actions—such as manipulating objects, changing viewpoints, or probing the environment—to resolve ambiguities, learn object properties, and refine multimodal representations~\cite{lynch2023interactive, queralta2020collaborative, novkovic2020object}. By leveraging exploratory actions, it allows EmAI systems to enhance or extend their perception abilities in object recognition~\cite{novkovic2020object}, scene understanding~\cite{hu2020interact}, or manipulation in dynamic and unstructured environments~\cite{lynch2023interactive}, \new{among others}. In robotic manipulation, interactive perception gathers data through exploration to identify potential actionable regions of objects and understand their functional possibilities. This process, \textit{a.k.a.} affordance learning, further benefits EmAI systems by guiding and optimizing their future interactions.

With affordance learning, action plans \new{can be generated} through two key approaches: supervised learning from human demonstrations~\cite{Luo2023Leverage, Wang2017Binge} and reinforcement learning from robotic trial-and-error interactions~\cite{graves2022affordance, Goff2019Building}. Pioneering approaches like Where2Act~\cite{mo2021where2act} enable robots to identify the most effective interaction strategies for various parts of an object, and Where2Explore~\cite{ning2024where2explore} generalize affordance knowledge to similar object parts, enabling robots to adapt to unseen objects with limited prior experience. Additionally, by integrating reinforcement learning, RLAfford~\cite{geng2023rlafford} facilitates end-to-end affordance learning, enabling robots to \new{seamlessly adapt} to a wide range of manipulation tasks.

\subsection{Low-level Actuation}
Low-level actuation is a fundamental component of EmAI systems that leverages various action control policies to determine real-time motor control based on perceptions~\cite{molchanov2019sim}. 
In this section, we explore the process of low-level actuation by dividing it into two core phases: control policy representation and control policy learning. The phase of \textit{\textbf{policy representation}} \new{presents} the framework for encoding robotic behaviors, ensuring that policies are expressive enough to capture intricate actions while remaining computationally efficient and adaptable to diverse scenarios. Building on this foundation, the \textit{\textbf{policy learning}} phase focuses on how robots select and optimize these behaviors through advanced algorithms including reinforcement learning, imitation learning, and hybrid strategies. Together, these two phases form a cohesive framework that equips robots with the ability to act autonomously and achieve predefined objectives. Table~\ref{table1} summarizes representative low-level control policies.

\begin{table*}[ht!]
\centering
\captionsetup{font=normalsize} 
\caption{Summary of Low-level Control Policies.}
\vspace{-0.25em}
\resizebox{\linewidth}{!}{%
\begin{tabular}{@{}p{4.7cm}p{4.7cm}p{9.7cm}@{}}
\toprule
\textbf{Task}                     & \textbf{Type}                        & \textbf{Research references} \\ \midrule

\multirow{3}{*}{Policy Representation} 
                                  & Explicit Policy                      & Mixture of Gaussians~\cite{mandlekar2021matters}, Categorical~\cite{shafiullah2022behavior}, DDPG~\cite{lillicrap2015continuous}, TD3~\cite{woo2020real}, SVG~\cite{heess2015learning}, REINFORCE~\cite{zhang2021sample}
                                  \\ \cmidrule(l){2-3}
                                  & Implicit Policy                      &
                                  EBMs~\cite{du2019implicit, Du2019ModelBP, boney2020regularizing}, Implicit BC~\cite{florence2022implicit}, Soft Q-Learning~\cite{haarnoja2017reinforcement}, IDAC~\cite{yue2020implicit}, Energy-Based Concept Models~\cite{mordatch2018concept},
                                  EBIL~\cite{liu2020energy}, ContactNets~\cite{pfrommer2021contactnets} \\ \cmidrule(l){2-3}
                                  & Diffusion Policy                     & XSkill~\cite{xu2023xskill},
                                  NoMaD~\cite{sridhar2024nomad}, Decision Diffuser~\cite{ajay2022conditional}, DALL-E-Bot~\cite{kapelyukh2023dall}, 
                                  ChainedDiffuser~\cite{xian2023chaineddiffuser},
                                  GSC~\cite{mishra2023generative}, HDP~\cite{ma2024hierarchical}, Equibot~\cite{yang2024equibot} \\ \midrule
\multirow{3}{*}{Policy Learning}  & Reinforcement Learning                         
                                  &Visual dexterity~\cite{chen2023visual}, DAPG~\cite{rajeswaran2017learning}, DDPGfD~\cite{vecerik2017leveraging}, ViSkill~\cite{huang2023value},
                                  Soft Actor-Critic~\cite{haarnoja2018soft}, SAM-RL~\cite{lv2023sam}, MT-Opt~\cite{kalashnikov2021mt} \\ \cmidrule(l){2-3}
                                  & Imitation Learning                             &GAIL~\cite{ho2016generative}, ValueDice~\cite{kostrikov2019imitation}, DeepMimic~\cite{peng2018deepmimic}, AMP~\cite{peng2021amp}, MPI~\cite{zeng2024learning}, 
                                  Vid2Robot~\cite{jain2024vid2robot}, Ag2Manip~\cite{li2024ag2manip}\\ \cmidrule(l){2-3}
                                  & RL \& IL Combination                           & AC-SSIL~\cite{liu2024surgical}, Guided Policy Search~\cite{levine2013guided}, Reward Shaping~\cite{hu2020learning}, 
                                  UniDexGrasp~\cite{xu2023unidexgrasp} \\ \bottomrule
\end{tabular}}
\vspace{-1em}
\label{table1}
\end{table*}

\subsubsection{Policy Representation}
Three types of policy representations \new{are} widely used in current EmAI systems. I) \textbf{\textit{Explicit policies}}~\cite{mandlekar2021matters, shafiullah2022behavior} is a common approach in behavior cloning~\cite{torabi2018behavioral} that maps directly from observations to actions. \new{This approach} can be viewed as a regression task designed to learn a mapping function, typically implemented via neural networks, to reproduce the behavior of the demonstrators. According to the output actions, deterministic policies~\cite{lillicrap2015continuous, kalashnikov2018scalable, woo2020real} provide fixed options, while stochastic policies~\cite{heess2015learning, liang2024rapid} encourage exploration by introducing randomness or initializing with high entropy. While straightforward to implement, explicit policies struggle with complex tasks due to insufficient expressive capability. II) \textbf{\textit{Implicit policies}}~\cite{florence2022implicit, wu2020spatial}, such as Energy-Based Models~\cite{du2019implicit, Du2019ModelBP, boney2020regularizing}, indirectly represent the policy through an energy function that defines the preference distribution in the state-action space, making them more suitable for multimodal distribution tasks. III) \textbf{\textit{Diffusion policies}}, inspired by Denoising Diffusion Probabilistic Models (DDPMs)~\cite{ho2020denoising}, create a policy \new{using} the conditional generative model~\cite{ajay2022conditional} and generate actions by refining noise, offering multimodal, expressive representations with applications in offline reinforcement learning~\cite{wang2022diffusion} and vision-based manipulation~\cite{chi2023diffusion}. While promising, diffusion policies require further optimization to improve sampling efficiency and inference speed in complex action spaces.

\subsubsection{Policy Learning}
Policy representation serves as the foundation for policy learning, enabling the encoding of behaviors that robots can execute. Based on policy representations, various learning algorithms, such as reinforcement learning~\cite{pateria2021hierarchical, zhang2022adjacency, hu2024transforming}, imitation learning~\cite{song2018multi, torabi2018behavioral, ho2016generative}, and hybrid approaches combining the two~\cite{levine2013guided, hu2020learning}, are employed to iteratively adjust and optimize the policy.

\textit{\textbf{Reinforcement Learning}} (RL) methods aim to optimize policies through trial-and-error interactions with the environment, guided by a scalar reward signal that encodes task performance. Recent RL algorithms can be broadly categorized into value-based, policy-based, and actor-critic approaches. Value-based methods, such as Deep Q-Networks (DQN)~\cite{huang2020deep}, estimate action values using temporal difference learning, enabling discrete action selection based on the greedy policy. In contrast, policy-based methods, like Proximal Policy Optimization (PPO)~\cite{schulman2017proximal} and Trust Region Policy Optimization (TRPO)~\cite{schulman2015trust}, directly optimize the policy in continuous action spaces by minimizing divergence between successive policies to ensure stable learning. Actor-critic methods~\cite{fujimoto2018addressing} combine these paradigms, where an actor (policy) generates actions, and a critic (value function) evaluates them, providing gradients to optimize the policy iteratively. However, RL algorithms often suffer from challenges such as sampling inefficiency, high exploration costs, and instability, which \new{limit} their practical applications in complex environments.

\textit{\textbf{Imitation learning}} (IL) provides an alternative by directly learning from expert demonstrations. These algorithms bypass the need for explicit reward functions, training policies to replicate expert behaviors using supervised learning techniques. Methods such as behavioral cloning~\cite{florence2022implicit, ly2020learning} minimize the discrepancy between predicted and expert actions, while inverse reinforcement learning~\cite{arora2021survey, ab2020inverse} aims to infer the underlying reward function from demonstrations. Although IL accelerates learning by reducing exploration, it struggles with covariate shift, where small errors during policy execution compound over time, \new{which limits} generalization in novel scenarios~\cite{ozalp2024advancements}.

\textit{\textbf{A hybrid approach that combines RL and IL}} leverages the strengths of both. Typically, policies are initialized with expert demonstrations to accelerate training using IL, and then refined by RL to adapt to dynamic environments~\cite{lu2023imitation, ozalp2024advancements, hua2021learning}. 
\new{Additionally, methods} such as Guided Policy Search~\cite{levine2013guided} or Reward Shaping~\cite{hu2020learning} \new{integrate} demonstration data into the RL process, enhancing both data efficiency and generalization. 
These combinations bridge the gap between the efficiency of the IL and the adaptability of the RL, improving the low-level actuation capability to execute actions in real-world scenarios.

\subsection{High-level Planning} 
Low-level actuation can only meet the needs of simple, reactive tasks, but it struggles to handle the complexity of planning long-horizon tasks with multiple sub-tasks. To address this limitation, high-level planning algorithms have been developed. \textit{\textbf{Classic planning algorithms}}, such as \textit{A*} algorithm~\cite{yao2010path, zhang2014multiple, tang2021geometric, erke2020improved}, Dijkstra's algorithm~\cite{javaid2013understanding, fan2010improvement, rachmawati2020analysis}, and the probabilistic roadmap (PRM) approaches~\cite{kavraki1996probabilistic, kavraki1998analysis, geraerts2004comparative, latombe1998probabilistic}, serve as foundational methods in this domain. Despite their significant influence and effectiveness in structured environments, these algorithms encounter substantial challenges in real-world scenarios, particularly in high-dimensional state spaces and under conditions of partial observability. Recent research \new{has explored the use of} LLMs as high-level planners in embodied systems, bridging cognitive reasoning and physical task execution by translating abstract instructions into actionable robotic tasks~\cite{yang2024embodied, ma2024survey}. Table~\ref{table2} summarizes representative approaches of different high-level planning methods.

\textit{\textbf{LLM-based task planners}} typically break down high-level goals into a sequence of executable subtasks~\cite{Birr2024AutoGPT+P:, Yang2024Text2Reaction, Li2022Hierarchical}. There are \new{generally} two main paradigms: \textit{\textbf{code-based planners}} and \textit{\textbf{language-based planners}}, as shown in the Figure~\ref{fig3}. Code-based planners~\cite{vemprala2023chatgpt, liang2023code, gu2024conceptgraphs} operate by selecting from a pre-defined set of modular skills or functions, invoking them via APIs to execute tasks step by step. They excel in highly controlled workflows requiring safety and reliability, where predefined APIs or modules can ensure predictable outcomes. Representative code-based planners such as the DEPS framework~\cite{wang2023describe} further emphasize ``self-explanation'' to better exploit the capabilities of LLMs, where an LLM generates plans, explains failures, and uses environmental feedback to aid in re-planning. In addition, by transforming observation sequences into 3D scene graphs using visual language models, ConceptGraphs~\cite{gu2024conceptgraphs} helps the LLM reason about spatial and semantic relationships in task planning. These methods benefit from the deterministic nature of pre-programmed functions, reducing ambiguity in execution. However, their dependency on predefined skills makes them less adaptable to unexpected changes or tasks outside the pre-programmed domain.
\vspace{5pt}

\noindent
\begin{minipage}[t]{0.50\textwidth}
    \hspace{12pt} Language-based planners~\cite{huang2022inner, song2023llm, yao2022react} show \new{greater} flexibility without using predefined functions. In open-ended domains like creative problem-solving, customer support, or planning in uncertain environments, language-based planners can use contextual cues to refine and adjust their instructions in real-time. Additionally, these methods dynamically integrate diverse feedback—ranging from success indicators to human inputs—without necessitating additional training for the LLMs. \new{Such} adaptive feedback and re-planning mechanisms~\cite{li2023interactive, wang2023describe, wang2024safe} enable systems to recover from unexpected states, offering more flexibility to novel environments. This adaptability, however, can sometimes come at the cost of consistency or precision~\cite{zeng2023large}. Moreover, by integrating MLLMs' capabilities, EmAI systems like Socratic models~\cite{zeng2022socratic} effectively translate non-language inputs into unified language descriptions through multimodal informed prompting. This approach not only streamlines information exchange across different modalities through MLLMs but also enhances robot perception and planning tasks~\cite{zheng2024planagent, zeng2023large, wang2024large}. 
\end{minipage}\hfill
\vspace{7pt}
\begin{minipage}[t]{0.47\textwidth}
    \vspace{-17pt}
    \centering
    \includegraphics[width=\textwidth]{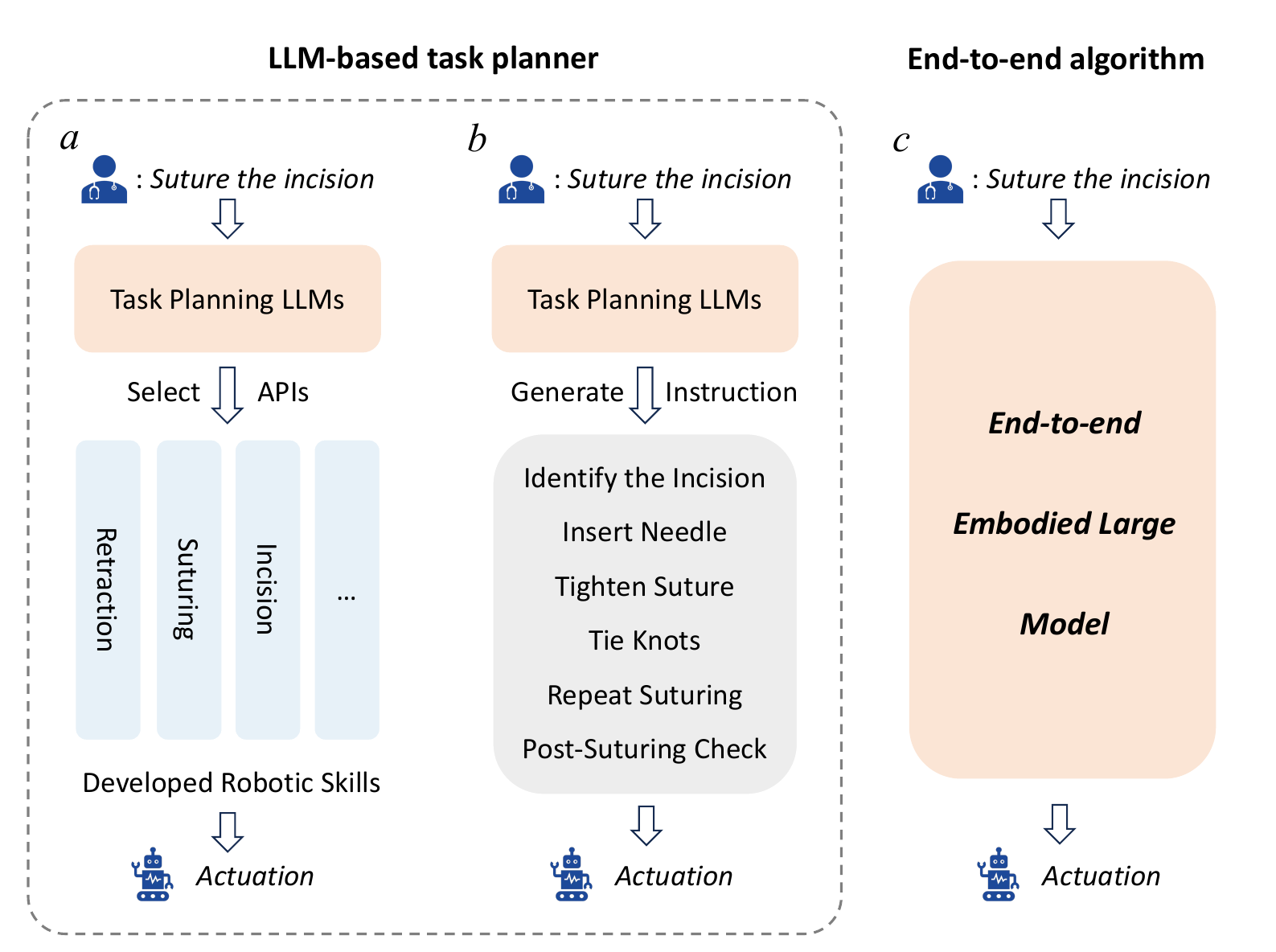}
    \captionsetup{skip=5pt} 
    \captionof{figure}{Three main approaches of high-level planning, using surgical robots as examples. (a) Code-based LLM task planners select and execute pre-developed robotic skills (e.g., retraction, suturing, incision) in actuation. (b) Language-based LLM task planners generate detailed task-specific instructions (e.g., identify the incision, insert needle, tie knots) for dynamic planning and execution. (c) End-to-end embodied large models directly integrate planning and execution in a single model.}
    \label{fig3}
    \vspace{-1em}
\end{minipage}
\vspace{7pt}

The high-level reasoning capabilities of LLMs have been further enhanced by integrating causal inference techniques~\cite{liu2024large}, transforming their planning ability from mere prediction to more logical and explanatory processes. This line of research encompasses prompt-based interventions, such as Chain-of-Thoughts~\cite{wei2022chain, madaan2023makes}, Tree-of-Thoughts ~\cite{yao2024tree}, and Graph-of-Thoughts~\cite{besta2024graph}, as well as interventions targeting inner LLM components~\cite{stolfo2023mechanistic} and causal graph abstraction~\cite{taori2023alpaca, wu2024interpretability}. Furthermore, Reinforcement Learning from Human Feedback (RLHF) has been pivotal in fine-tuning LLMs by leveraging human evaluations to guide model behavior~\cite{bai2022training, kak2024embodied}, enabling EmAI to behave more like humans and demonstrate improved explainability.

\textit{\textbf{Planning with end-to-end Embodied Large Models.}} These algorithms stand out by directly mapping high-level instructions to low-level actions, seamlessly integrating perception, planning, and control into a unified system. Recent research shows that these systems leverage deep reinforcement learning and imitation learning to streamline planning and decision-making in complex environments, often surpassing modular pipelines in adaptability and robustness~\cite{Francis2021Core, Chen2023End-to-end}. Notable frameworks such as SayCan~\cite{ahn2022can}, PaLM-E~\cite{driess2023palm}, and EmbodiedGPT~\cite{mu2024embodiedgpt} are designed to combine vision encoder embeddings with planning data from LLMs, directly informing the robotic policies for immediate action. However, training embodied end-to-end systems typically requires large-scale datasets. To address this need, simulation data is widely used for its efficiency and safety, though challenges arise from discrepancies in physics, sensors, and real-world complexity. Strategies such as domain randomization~\cite{Chen2021Understanding}, domain adaptation~\cite{Exarchos2021Policy}, and hybrid approaches~\cite{Valdivia2024Safe} have been developed to enhance adaptability and bridge the sim-to-real gap, improving real-world performance. In addition, it is also important to design benchmarks to evaluate the embodied decision-making capability among different LLMs~\cite{li2024embodied}.

\begin{table*}[ht!]
\centering
\fontsize{19}{38}\selectfont
\captionsetup{font=normalsize} 
\caption{Summary of High-level Planning Approaches.}
\resizebox{\linewidth}{!}{%
\begin{tabular}{@{}p{8cm} p{6cm} p{22cm}@{}}
\toprule
\textbf{Methods}                     & \multicolumn{2}{c}{\textbf{Research references}} \\ \midrule
\textbf{Classic planning algorithm}           & \multicolumn{2}{c}{\textit{A*} algorithm-based planners~\cite{tang2021geometric, erke2020improved}, Dijkstra's algorithm~\cite{fan2010improvement}, PRM planners~\cite{kavraki1996probabilistic, latombe1998probabilistic}} \\ \midrule
\multirow{3}{*}{\textbf{LLM-based planner}}   & Code-based:  & Code-as-Policies~\cite{liang2023code}, ProgPrompt~\cite{singh2023progprompt}, Instuct2Act~\cite{huang2023instruct2act}, ChatGPT for Robotics~\cite{vemprala2023chatgpt}, Chain od Code~\cite{li2023chain}, DEPS~\cite{wang2023describe}, ConceptGraphs~\cite{gu2024conceptgraphs} \\ \cmidrule{2-3}
                                      & Language-based: &  Natural Language as Policies~\cite{mikami2024natural}, 
                                      Inner Monologue~\cite{huang2022inner}, 
                                      Statler~\cite{yoneda2024statler},
                                      LLM-Planner~\cite{song2023llm}, ReAct~\cite{yao2022react}, Socratic Models~\cite{zeng2022socratic} \\ \midrule
\textbf{End-to-end algorithm}                 & \multicolumn{2}{c}{\rule{0pt}{42pt}\parbox{28cm}{\centering
EmBERT~\cite{suglia2021embodied}, SayCan~\cite{ahn2022can}, PaLM-E~\cite{driess2023palm}, MS-PCD~\cite{zhao2024see}, Zero-Shot Trajectory~\cite{kwon2024language}, \\ EmbodiedGPT~\cite{mu2024embodiedgpt}, Planning as Inpainting~\cite{yang2024planning}, LID~\cite{li2022pre}, (SL)³~\cite{sharma2021skill}, LACMA~\cite{yang2023lacma}, LEO~\cite{huang2023embodied}
}\rule[-28pt]{0pt}{28pt}} \\ \bottomrule
\end{tabular}}
\label{table2}
\end{table*}

\subsection{Memory Processing}
Memory serves as a repository for past experiences and knowledge, allowing systems to learn from historical data, adapt to new situations, and make informed decisions based on accumulated insights~\cite{wang2024karma}. Memory in EmAI systems is typically divided into short-term memory and long-term memory, each serving complementary functions.

\textit{\textbf{Short-term memory}} employs mechanisms such as in-context prompts~\cite{wang2023exploring, ahn2022can, rana2023sayplan} and latent embeddings~\cite{liu2024visual, zhu2023minigpt} within LLMs to manage immediate data needs during interactions. This type of memory often holds data from ongoing interactions and is critical in settings involving dialogues and environmental feedback. For instance, chatbots maintain conversation histories to facilitate ongoing exchanges, while EmAI systems may use textual representations of environmental feedback as a form of short-term memory, aiding in immediate reasoning tasks~\cite{yao2022react}. This allows EmAI systems to temporarily prioritize new over old information and adapt to new situations with recently learned knowledge. 

\textit{\textbf{Long-term memory}} serves as a foundational component by storing vital, factual knowledge that influences EmAI systems' actions and their understanding of the world~\cite{gao2024empowering}. The integration of long-term memory allows LLMs to leverage past experiences during inference, thereby enhancing their self-evolution capabilities and proficiency in handling complex tasks~\cite{wang2024augmenting, hatalis2023memory, liu2023think}.
Long-term memory is structured into internal and external systems: the internal memory is embedded within the AI model's own architecture through model weights, enabling swift, direct zero-shot application of learned information~\cite{touvron2023llama, anil2023gemini}, while the external memory, stored in separate databases or knowledge graphs, requires active retrieval and integration for usage~\cite{wiest2024privacy, zhong2024memorybank, hu2023chatdb,liang2024genetic, lyu2024gp, cinquin2024chip}. 
To stay current, long-term memory stored in models can be dynamically updated through fine-tuning techniques, such as supervised fine-tuning (SFT)~\cite{lu2023instag}, instruction fine-tuning (IFT)~\cite{ouyang2022training}, and parameter-efficient fine-tuning (e.g., LoRA)~\cite{dettmers2023qlora, hu2021lora}, while the external memory is updated by directly improving external databases.

\subsection{Synergistic Integration}
The aforementioned four key functionalities are often initially developed independently but must be effectively integrated to construct a comprehensive EmAI system. Integration approaches such as MemoRAG~\cite{qian2024memorag} and Reflexion~\cite{shinn2024reflexion} enhance high-level planning processes by retrieving relevant information~\cite{lewis2020retrieval, jiang2023active, yu2024rankrag} or summarizing past experiences~\cite{jiang2024koma, dou2024reflection} from memory modules. These methods improve adaptability to dynamic environments and enable more rational decision-making. Closed-loop approaches, including RoboGolf~\cite{Zhou2024RoboGolf:}, LyRN~\cite{zhuang2020lyrn}, and AlphaBlock~\cite{jin2023alphablock}, integrate perception and actuation modules, leveraging feedback to refine observations and dynamically update control signals. This integration facilitates precise action adjustments and supports effective multi-step planning. \new{Moreover}, active and interactive perception systems~\cite{novkovic2020object, zhou2023enhancing, zhao2023chat} take this a step further by engaging in real-time interactions to explore object properties, update environmental contexts, and refine decisions based on immediate outcomes. The resulting actions and observations are stored in the EmAI memory, which can be used to construct multimodal knowledge graphs integrating physical properties, concepts, affordances, and intentions for future use~\cite{song2024scene}.
The modern AI alignment approaches that combine modules~\cite{liu2023combining, ji2023ai} and foundational models that encompass all functionalities of these modules~\cite{chen2023towards, zhou2024embodied} are also being widely researched as promising areas~\cite{xiang2024language, xu2024large}. However, there is still a lack of a highly compatible, efficient, and effective unified architecture capable of integrating various developed modules. Achieving alignment and seamless integration among these modules, while minimizing development (e.g., fine-tuning), remains an open challenge.

\section{Applications of Embodied AI in healthcare}
\label{section3}

This section presents healthcare applications and products of EmAI systems across four key domains: \textit{\textbf{Clinical Intervention}}, \textit{\textbf{Daily Care \& Companionship}}, \textit{\textbf{Infrastructure Support}}, and \textit{\textbf{Biomedical Research}}. 

\begin{itemize}
    \item \textit{\textbf{Clinical Intervention}} involves targeted actions to treat or manage medical conditions, and EmAI systems can provide robot-assisted diagnosis~\cite{salcudean2022robot}, precision intervention~\cite{loftus2023artificial}, and personalized postoperative rehabilitation~\cite{sumner2023artificial}.

    \item \textit{\textbf{Daily Care \& Companionship}} relies on AI-driven robots to assist the elderly and individuals with disabilities by monitoring health, aiding mobility, and providing emotional support, improving quality of life and reducing caregiver burden.

    \item \textit{\textbf{Infrastructure Support}} benefits from EmAI systems that improve efficiency and safety through tasks like emergency response, pharmaceutical distribution, environmental disinfection, and patient transport.

    \item \textit{\textbf{Biomedical Research}} leverages EmAI systems to accelerate discoveries by automating experiments, conducting high-throughput analyses, and interpreting complex biological data.
\end{itemize}

\subsection{Clinical Intervention}
\begin{figure}[t]
    \centering
    \includegraphics[width=\linewidth]{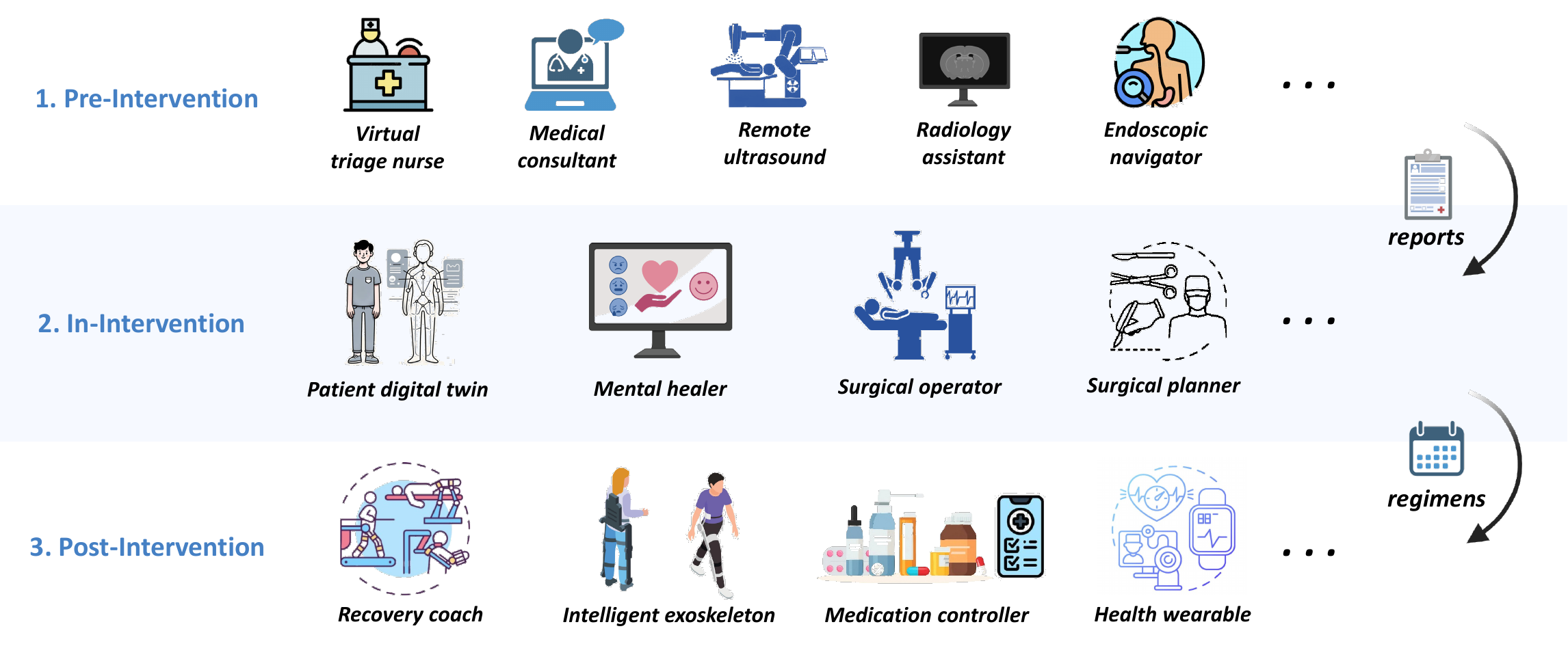}
    \caption{Embodied AI applications play critical roles in clinical intervention across the pre-intervention, in-intervention, and post-intervention phases. These systems enhance precision by performing essential functions such as diagnosis and planning, supporting surgical and therapeutic procedures, and facilitating recovery and health management. These EmAI systems can provide prior information or execute follow-up tasks for others, highlighting their potential for integrated development. By unifying these capabilities within a single embodiment, it is possible to create a comprehensive EmAI system for clinical intervention, with the capacity to significantly improve patient care across the entire clinical intervention spectrum.}
    \label{fig4}
    \vspace{-0.5em}
\end{figure}

EmAI systems have been extensively applied in clinical interventions, spanning the pre-intervention, in-intervention, and post-intervention phases~\cite{salcudean2022robot, loftus2023artificial, sumner2023artificial}. We outline their primary roles in this section, as shown in Figure~\ref{fig4}.

\subsubsection{Pre-Intervention Stage}
Recent improvements in EmAI-related technologies for pre-intervention diagnostics and assessments are shaping a new AI-clinician collaboration in the intelligent hospital~\cite{hunter2022role, chen2024ai, fan2020investigating}. EmAI plays various roles in this context, reducing clinicians’ workloads and accelerating diagnostic workflows. 

\textit{\textbf{Virtual triage nurse.}} In modern smart healthcare systems, EmAI-based virtual triage nurses \new{are replacing} human nurses and play a pivotal role in streamlining patient management by directing individuals to the most appropriate clinical departments. These EmAI systems analyze patient-reported symptoms and \new{rank} departments based on symptom descriptions~\cite{Fernandes2020Clinical, yang2024evolution, menshawi2025novel}. 
Beyond symptom-based sorting, advanced triage systems integrate wearable health data and EHR to provide a holistic assessment of conditions~\cite{Dinh-Le2019Wearable, polley2021wearable, salman2021review}. In emergency care settings, these systems can even predict patient outcomes and recommend intervention pathways, significantly shortening response times during critical situations~\cite{nino2020coupling, shibu2024cloud, kipourgos2022artificial,chen2024artificial}. As healthcare systems become more interconnected, intelligent triage systems are increasingly serving as an efficient tool \new{for} future infrastructure~\cite{salman2021review, weisberg2020first}, enabling seamless coordination between primary care, specialist consultations, and hospital admissions.

\textit{\textbf{Interactive medical consultant.}} Recently, some LLM-based chatbots, such as DISC-MedLLM~\cite{bao2023disc} and HealAI \cite{goyal2024healai}, \new{have been developed} to provide instant, reliable, and context-specific responses to medical inquiries, helping patients better understand their symptoms, treatment options, or follow-up care~\cite{hwang2020implementation, hwang2020implementation, zhang2025revolutionizing}. They can also explain medical conditions, offer personalized recommendations~\cite{srivastava2020automatized, dharwadkar2018medical}, and explain radiology reports~\cite{wang2023chatcad}. By bridging the gap between patients and complex medical knowledge, interactive medical chatbots not only empower individuals to make informed decisions but also reduce the workload of healthcare professionals~\cite{chakraborty2022ai, hsu2022medical}. With advancements in LLM reasoning~\cite{yang2024drhouse}, these systems are becoming an essential component of patient-centered smart healthcare solutions. Beyond answering medical questions, these chatbots can also guide patients through administrative processes, such as booking appointments~\cite{khamaj2025ai, srivastava2023machine} or managing prescriptions~\cite{gudala2022benefits, roca2020virtual}, thereby easing the burden on healthcare staff~\cite{nivedhitha2024conversational, milne2020effectiveness}. With their ability to operate around the clock, these systems foster greater accessibility and trust in the medical process.

\textit{\textbf{Imaging analyst.}} Another representative area is robot-assisted medical imaging, which not only improves the accuracy of diagnostic processes but also expands the capabilities of medical imaging in complex anatomical assessments. \new{Many studies have} focused on medical image analysis using AI techniques~\cite{suganyadevi2022review,kaur2023artificial,wang2024lkm}, but to support robotic surgery and preoperative robotic diagnosis requires additional requirements including real-time processing~\cite{park2020deep}, 3D spatial understanding~\cite{haidegger2022robot}, and safety compliance~\cite{mennella2024ethical} beyond high precision. Among medical imaging technologies including ultrasonography, radiology, and endoscopy, EmAI enhances their capabilities in different ways. 

For ultrasonography, \new{which} is portable, real-time, non-invasive, and relying on the synchronization of diagnostic and procedural operations, EmAI can serve as a remote assistant to help clinicians conduct remote ultrasound diagnosis~\cite{hidalgo2023current} and protect themselves from risks against epidemics~\cite{zemmar2020rise}. For radiology, EmAI can also help clinicians in lesion localization~\cite{salcudean2022robot, iftikhar2024artificial, Ferrari2023Autonomous}, surgical planning~\cite{Ujiie2024Developing, Park2023Patient-specific, khizir2024diagnosis}, intraoperative navigation~\cite{Li2021A, Müller2020Augmented} with a better understanding of the raw imaging and attached report from X-Ray, Computed Tomography (CT), and Magnetic Resonance Imaging (MRI). For endoscopy, EmAI can process endoscopic video feeds in real-time to identify abnormalities~\cite{okagawa2022artificial, wu2022real}, such as polyps or early-stage tumors, reduce operator fatigue, enable precise navigation via path optimization~\cite{kim2020autonomously, chen2024toward}, and assist in procedural tasks like polypectomy with adaptive, precision-controlled robotic movements~\cite{ma2019autonomous, barua2021artificial}.

\begin{itemize}

\item \textit{\textbf{Remote ultrasound.}} A major benefit of robotic ultrasound is its ability to be remotely operated, facilitating medical diagnosis in remote areas and reducing the healthcare gap between rural and urban communities~\cite{hidalgo2023current}. One of the most representative remote ultrasound diagnosis methods is image-based visual servoing algorithms~\cite{akbari2021robot, wang2024autonomous, von2021medical}, which can adjust the probe's orientation and position remotely in real time, dynamically refining the movements and force of robotic arms~\cite{9336261, 9610119, pore2021safe} in response to changes of patient anatomy, or procedural requirements of diagnostics. Complementary methods such as medical image registration~\cite{fu2020deep} have also been proposed to assist robotic ultrasound scanning, improving positioning accuracy~\cite{jiang2023skeleton, haskins2020deep}, performing motion compensation~\cite{suligoj2021robust, jiang2022precise}, and enabling real-time monitoring~\cite{zhang2020artificial, xu2024transforming}.

\item \textit{\textbf{Guardian against epidemics.}} Another potential usage of robotic medical imaging systems is to use their teleoperated nature as a protective barrier, physically separating healthcare workers from infected patients~\cite{zemmar2020rise}. This approach helps to address widespread concerns about exposure to infection and allows clinicians to focus on providing high quality care without compromising their safety or that of their patients. Tele-operated robotic lung ultrasound system has gained attention~\cite{tsumura2021tele, feizi2021robotics} that enabled remote assessment of lung conditions, effectively reducing the risk of viral transmission. Similarly, AI-Corona, a radiologist-assistant framework, enables COVID-19 diagnosis through chest CT scans with faster and more precise assessments while minimizing patient-clinician interaction~\cite{yousefzadeh2021ai}. Such advancements improve the efficiency and safety of healthcare delivery during pandemics.

\item \textit{\textbf{Radiology assistant.}} Interventional Radiology (IR)~\cite{lanza2023robotics} focuses on using image-guided, minimally invasive techniques to perform diagnostic and therapeutic procedures. The integration of EmAI is further advancing IR by improving procedural accuracy, safety, and operator control in complex medical interventions. For example, robotic systems have applied CT images to improve the accuracy of localization of lesion areas such as pulmonary nodules~\cite{liu2023robotic} and navigated placement of thoracolumbar pedicle screws~\cite{wang2021robotic} in both preoperative diagnosis and intraoperative procedures. Similarly, robotic systems leveraging X-ray imaging have enhanced cardiac interventions by enabling precise catheter navigation in endovascular catheterization~\cite{omisore2021automatic, gunduz2021robotic}. MRI-guided robotic interventions have facilitated resection margins targeting in procedures like tumor ablation, providing superior visualization of real-time tissue changes during operations to protect critical structures~\cite{fang2021soft, huang2023mri, su2022state}.

\item \textit{\textbf{Endoscopic navigator.}} The integration of EmAI in endoscopic procedures has significantly enhanced the precision and efficiency of minimally invasive diagnostics and treatments by providing real-time guidance for device positioning and lesion targeting~\cite{li2018robotic,zinchenko2021autonomous}. For example, AI-powered 3D mapping techniques reconstruct anatomical structures in real time, enabling clinicians to navigate complex regions such as the gastrointestinal tract and respiratory pathways with greater accuracy~\cite{bovskoski2021robotics}. Robotic-assisted endoscopy further \new{utilizes} automated control systems that adapt dynamically to patient-specific anatomy and procedural demands~\cite{ng2024simultaneous, jones2022combined}, \new{which improves} precision in tasks such as biopsy, polyp removal, and other targeted interventions. Furthermore, these systems incorporate predictive analytics to refine surgical pathways, reduce tissue trauma, and streamline procedural workflows, enhancing safety and operator confidence in endoscopic interventions~\cite{simsek2023future, feng2024toward}.

\end{itemize}

In summary, fully autonomous ultrasound system~\cite{su2024fully}, human-centric radiology assistant~\cite{calisto2021introduction}, and flexible robotic endoscope system~\cite{mo2022task} capable of navigating and adjusting diagnostic examinations with minimal human supervision have been developed, achieving a higher level of intelligence and autonomy. Compared to human-led examinations, \new{these systems} optimize workflows, improve patient care quality, and ensure a safer environment for all involved in the surgical process. \new{Additionally,} these systems can also function as a module within a more comprehensive EmAI system such as surgical robots, showcasing opportunities to expand their applications.

\subsubsection{In-Intervention Procedure}

The ongoing integration of EmAI systems into intervention procedures has catalyzed advancements in various fields, including surgical practice, mental health interventions, and beyond. Among these, surgical practice has emerged as the most extensively studied and developed application to date. These systems automate specific surgical tasks and provide critical intraoperative feedback, thereby improving both the execution of operations~\cite{marinho2021smartarm, wu2024surgicai, schorp2023self} and the analysis for surgical training~\cite{holm2023dynamic, innocent2024surgitrack, cui2024surgical}. A representative hierarchical EmAI system for automating operations is shown in Figure~\ref{fig5}.

Traditional robotic surgery has primarily focused on developing task-specific policies for surgical actions such as suturing~\cite{shademan2016supervised, pedram2017autonomous, zhong2019dual}, tissue manipulation~\cite{shin2019autonomous, tagliabue2020soft}, and gauze cutting~\cite{thananjeyan2017multilateral}. The rapid advancements in LLMs and VLAs \new{have further enhanced} the intelligence and versatility of \new{these systems}.

\textit{\textbf{Surgical planner.}} In conventional clinical settings, meticulous surgical planning is made by surgeons at the initial phase of surgery. With EmAI, this process can be \new{automated and streamlined}.
Emerging VLA models have been applied in this field to generate actionable surgical plans~\cite{li2024robonurse}. They allow robots to understand visual cues and instructions in natural language, enabling more intuitive and flexible task execution~\cite{zeng2023large, schmidgall2024gp}. In addition, some VLA models facilitate efficient imitation learning to learn from visual demonstrations to learn complex actions. By interpreting human demonstration videos, EmAI systems are able to recognize fine-grained visual concepts including instruments, verbs, and targets~\cite{yuan2023learning} and create actionable task plans~\cite{wang2024vlm} for robotic systems in surgical settings. 

Moreover, surgical planners powered by Vision-Language Models (VLMs)~\cite{li2025benchmark}, such as SurgicalGPT~\cite{seenivasan2023surgicalgpt} and LLaVA-Surg~\cite{li2024llava}, can benefit both practical and educational purposes by offering assistance and evaluation during surgical procedures. Through quantifying surgical performance by decomposing complex surgeries into discrete gestures, the impact of different surgical gestures on patient outcomes can be statistically analyzed and used to predict postoperative results through EmAI systems~\cite{ma2022surgical}. 
Similarly, other studies~\cite{yip2023artificial} \new{support the idea} that developing precise, computer-aided surgical planning can help improve patient outcomes. It is suggested that equipping the \textit{da Vinci system}~\cite{surgical2013vinci} with AI-based perceptual capabilities can enhance its ability to understand the operating scenarios, plan surgeries, and execute better operations~\cite{wu2024surgicai}. 

While the prior discussion centered around the planning of surgical skills into action-level details, within the contemporary medical landscape, there are complex multi-stage surgeries with long-term plans, dividing complex surgeries into multiple sessions for execution. Such surgeries often involve intricate coordination among surgical teams, precise management of patient health over extended periods, and adaptation to unforeseen intraoperative changes.
Compared with traditional surgical planning methods, EmAI can enhance multi-stage planning for complex surgeries by integrating various system modules, such as diagnostic tools~\cite{Ujiie2024Developing, Park2023Patient-specific, Qin2024Collaborative}, real-time monitoring systems~\cite{swarnakar2023artificial, dias2020innovative, cha2021deep}, and predictive analytics~\cite{shouval2014application, van2022critical} for long-term patient health management. These AI-driven modules facilitate the dynamic adaptation of clinical intervention schedules and strategies based on the assessment of patient conditions during successive surgeries~\cite{ma2022machine, wang2021operating}. While significant advances have been achieved in individual EmAI modules, seamlessly integrating them into a fully functional surgical planner with global cognition remains a critical focus for future innovation.
\vspace{7pt}

\noindent
\begin{minipage}[t]{0.44\textwidth}
    \hspace{12pt} \textit{\textbf{Surgical operator.}} Skillful operators are at the core of every surgical procedure, where precision and timely intervention are crucial. Challenges often arise due to the intricate nature of surgeries and the finite availability of expert operators, which can lead to bottlenecks in patient care. To enhance this, robotic surgery has been a promising trend to offer enhanced precision, dexterity, and minimally invasive approaches to common surgical procedures, leading to improved patient outcomes~\cite{sheetz2020trends}. In healthcare, a dedicated robot is often more practical and reliable and has been extensively studied in a variety of settings, \new{demonstrating} overall clinical benefits~\cite{muaddi2021clinical}. 
\end{minipage}\hfill
\vspace{7pt}
\begin{minipage}[t]{0.53\textwidth}
    \vspace{-10pt}
    \centering
    \includegraphics[width=\textwidth]{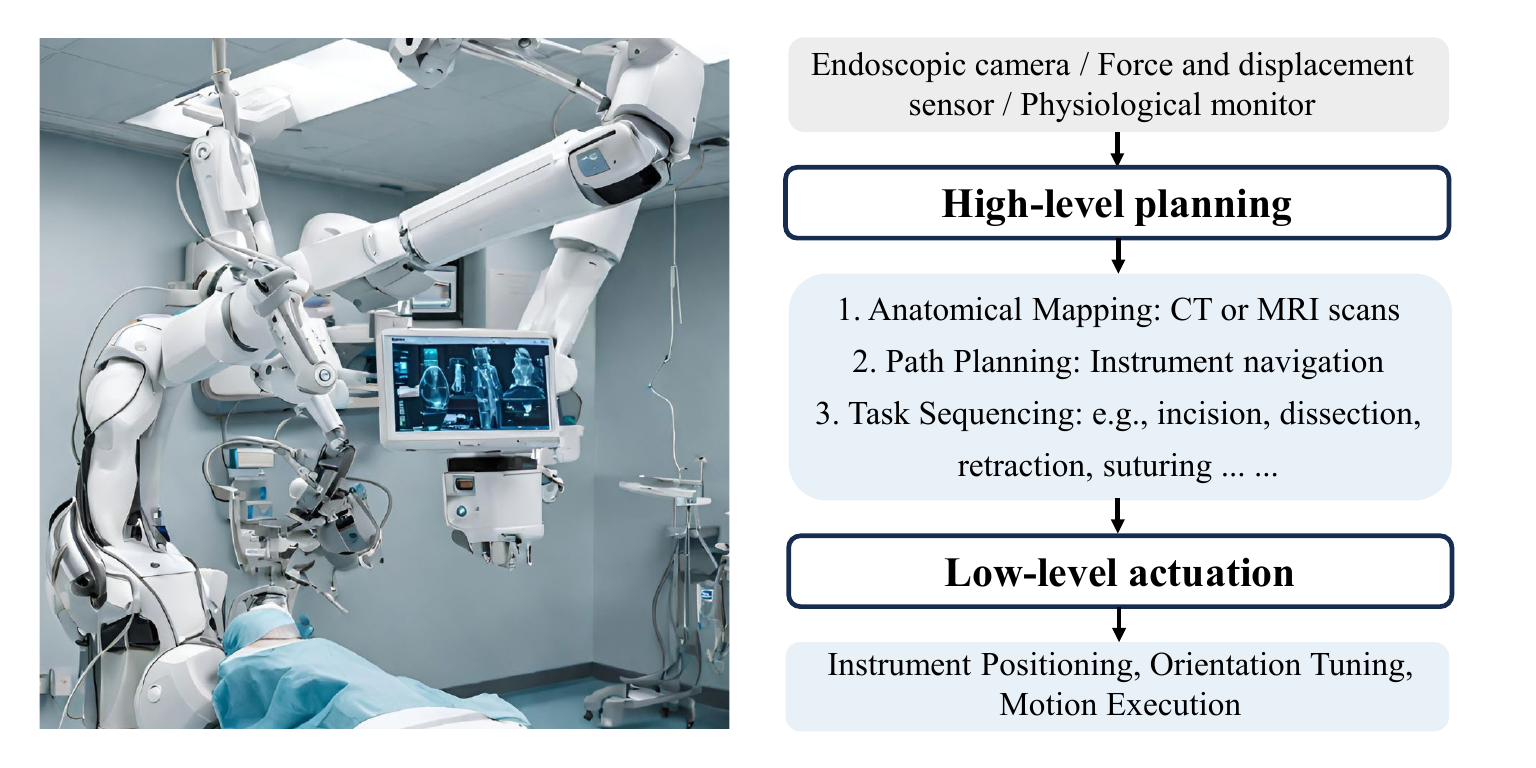}
    \captionsetup{skip=5pt} 
    \captionof{figure}{An EmAI system for surgical robotics: it combines strategic planning with precise actuation. This structure optimizes the execution of complex surgical tasks, from detailed pre-operative mappings to real-time adjustments during surgery.
    \textcolor{myColor}{\textbf{Source:} Reproduced from \cite{Cashman2024SurgicalRobotics}. Copyright © Boston Engineering 2024.}}
    \label{fig5}
    \vspace{-1em}
\end{minipage}
\vspace{7pt}

For example, in laparoscopic surgery, there are robotic systems autonomously performing intestinal anastomoses~\cite{saeidi2022autonomous} and rectal cancer resection~\cite{safiejko2021robotic} under human oversight. In retinal microsurgery, comprehensive EmAI systems~\cite{kim2024towards, iordachita2022robotic} have been developed to integrate real-time surgical object tracking, segmentation, and model predictive control for intraoperative navigation in delicate microsurgical environments \new{where} precision is decisive. 

For general surgical skills, the potential of robotics has also been widely demonstrated. Robots driven by deep learning and active sensing strategies \new{have been} trained to use needles through the HOUSTON algorithm~\cite{wilcox2022learning}, which can localize, grasp, and hand over unmodified surgical needles to complete precise manipulation. 
Their \new{exceptional} suturing skill that requires a stable and delicate control system has also been proved by the SmartArm robotic system for a neonatal chest surgery~\cite{marinho2021smartarm}. Similarly, for thread detection and interactive perception, real-time robotic approach~\cite{schorp2023self} \new{has been} presented for autonomously shortening the tail of a surgical suture using a self-supervised learning framework. In complex surgical scenes, multiple EmAI subsystems may collaborate to integrate multi-source perception and support robotic surgery, requiring a robust control system to direct and coordinate their respective perception-cognition-communication-action loops~\cite{de2024teamcollab}. Beyond visual perception, other modalities including kinematic data~\cite{long2021relational}, audio data~\cite{chen2024bootstrapping}, language instructions~\cite{xu2024transforming}, and tactile perception~\cite{liu2020embodied} are integrated to form a holistic view of real-world surgical environments.

\textit{\textbf{Intelligent surgeon-robot interface.}} Intuitive user interfaces and advanced responsive systems can significantly bridge the gap between surgeons and robotic tools, reducing the difficulty of surgeon-robot collaboration and facilitating communication and control as a ``translator". Two kinds of intuitive control and sensory feedback approaches were explored. A novel voice control interface~\cite{davila2024voice} for surgical robots using Whisper speech recognition technology allows surgeons to command robots verbally. Sensory gloves~\cite{borgioli2024sensory} that can translate natural hand movements to control robotic tools \new{aim} to provide the tactile feedback that is crucial during surgery but is often missing in robotic operations. Furthermore, human-in-the-loop learning systems~\cite{long2023human, liu2022robot} involve human interaction to guide and refine the learning process of AI models. These approaches ensure that EmAI is continually improved by incorporating human expertise and feedback for better decision making, which is particularly effective and intuitive in the knowledge-intensive surgical environment. An interactive system that allows surgeons to guide the robot's learning process during surgery can be more reliable by reducing the error rate, while at the same time being less fully automated.

\textit{\textbf{Surgical navigator.}} Intraoperative navigation systems significantly enhance surgical procedures by improving the precision of tool positioning, optimizing surgical pathways, providing real-time feedback, and reducing operative risks. These systems enable surgeons to execute complex surgeries with heightened accuracy and control, minimizing damage to surrounding healthy tissues and improving overall surgical outcomes. Emerging VLN technologies~\cite{park2023visual, huang2023visual} further enable robots to adapt to diverse surgical environments, respond to verbal commands, and navigate autonomously. Researchers have developed models that integrate spatial awareness and task-specific knowledge to enhance robots' responsiveness to natural language instructions~\cite{bieck2020language, jeong2024survey}. 

There are two core components of a surgical navigator, one is the ability to understand and contextualize visual scenes, and the other is safety-oriented path planning and obstacle avoidance. Surgical environments are crowded, and tasks require high precision, making it crucial for AI to correctly identify anatomical landmarks, surgical tools, and other visual cues. Advanced VLN models combine vision transformers and LLMs to enable contextual understanding from both visual and textual inputs~\cite{zeng2023large}. In addition, navigation systems must prioritize safety above all in a surgical setting. AI-driven robots need to navigate within confined spaces while avoiding obstacles, such as surgical instruments, medical staff, or sensitive patient tissues. Path-planning algorithms tailored to the surgical domain incorporate safety constraints and predictive models that anticipate potential obstructions~\cite{xu2023information, zhang2023novel}. Techniques such as depth sensing and 3D scene reconstruction are employed to enhance spatial awareness~\cite{bartholomew2024surgical, chen2024surgicalgs, yang20243d}, \new{allowing} robots to navigate autonomously while ensuring they remain within a safe operating range.

Embodied AI not only assists in the execution of surgical tasks but also enhances intraoperative decision-making through precise feedback and comprehensive analysis. 

\textit{\textbf{Real-time operation consultant.}} The modern surgical environment typically involves processing vast amounts of real-time information. Complex procedures generate extensive visual and contextual data, such as live video feeds, laparoscopic images, and robotic surgery footage, which must be rapidly interpreted under high-pressure conditions. Simultaneously, even experienced surgeons may encounter scenarios that exceed their expertise, and reliance on human judgment alone can lead to inefficiencies or errors. To bridge these gaps, AI-powered question-answering approach that is capable of interpreting surgical contexts and providing precise and in-time answers is an essential innovation. Some EmAI systems capable of answering questions based on visual data from surgical environments have emerged~\cite{seenivasan2022surgical, yuan2024advancing,antonio2024croma}. These functionalities are typically enabled by VQA and image captioning approaches, and recent advancements~\cite{li2024llava, jin2024surgical, schmidgall2024gp, he2024pitvqa, chen2024llm, wang2024surgical,hu2024ophclip} employing novel Vision-Language Pretraining (VLP) techniques in surgery-specific VQA and image captioning tasks have further enhanced the capabilities of EmAI systems, where pre-trained multimodal models are fine-tuned with surgery-specific VQA datasets.

\textit{\textbf{Surgical operation coach.}} The increasing complexity of surgical procedures, combined with the limited availability of experienced mentors for young trainees, has created significant challenges in providing adequate education for novice surgeons. This training gap \new{further} exacerbates the global shortage of skilled surgical professionals. Robotic surgical operation coaches, integrated with advanced EmAI algorithms, offer a promising solution to this pressing issue. Researchers have developed EmAI models capable of real-time recognition and prediction of surgical gestures and trajectories~\cite{shi2022recognition, weerasinghe2024multimodal, gazis2022surgical, shi2023haptic, feghoul2024mgrformer}. To achieve a comprehensive understanding of the surgical environment, it is crucial to undertake a multi-granular analysis of surgical activities. This includes long-term tasks, such as recognizing surgical phases and steps, as well as short-term tasks, such as segmenting surgical instruments and detecting atomic visual actions~\cite{valderrama2022towards}. Through these advancements, surgical operation coaches can analyze the intricate sequences of surgical activities, providing objective evaluations of surgical skills~\cite{10585847, zhang2024cwt, zheng2024transformer, quarez2024r, anastasiou2023keep, feghoul2023spatial}. These evaluations \new{provide} constructive feedback, enabling trainees to refine their techniques and accelerate their learning curve.

\textit{\textbf{Patient digital twin.}} A patient digital twin represents a detailed, dynamic model of a patient’s biological systems or a part thereof (e.g., anatomy), created using comprehensive and accurate medical data such as imaging studies, physiological measurements, and diagnostic results~\cite{chaby2022embodied}.
It is often combined with advanced visualization devices (e.g., VR/AR) to provide interactive observations, \new{allowing surgeons to} plan, simulate, and optimize surgical pathways based on them~\cite{inamura2023digital, ahmed2021potential}. AI-based digital twin creation technology facilitates the identification of surgical patterns~\cite{holm2023dynamic}, prediction of complications or procedural outcomes~\cite{ahmed2021potential}, and generation of medical reports~\cite{lin2022sgt, lin2023sgt++}. This technology is often employed to accelerate the learning curve of novice surgeons, helping them familiarize themselves with anatomical structures, surgical procedural contexts, and disease progression~\cite{ding2024digital, hagmann2021digital,wang2024twin,hu2024personalized,chen2021electrocardio}. Furthermore, multiple downstream applications have benefited from the surgery digitization process~\cite{shu2023twin}. Clinicians can practice surgical skills and explore human anatomy without relying on real anatomical models using digitalized surgical platforms~\cite{sun2023digital}, overcoming the limitation of high costs and scarce training samples. It also offers a digital and accurate platform for robots to train on, improving safety and reliability before deployment in actual surgical procedures~\cite{hein2024creating}. Another approach involves photorealistic surgical images~\cite{nwoye2024surgical} and videos~\cite{cho2024surgen} synthesis to benefit the training process of EmAI systems, mitigating challenges related to the high costs and ethical concerns of obtaining surgical data.

The developments in EmAI systems for operation and related supporting roles are reshaping surgical practices by enhancing efficiency and precision. These systems provide valuable products and analysis, crucial for both current surgical procedures and the training of future surgeons.
However, EmAI-based interventions encompass more than \new{just crucial surgical applications}; they also extend to interventional fields \new{such as} mental health therapy.

\textit{\textbf{Mental healer.}} EmAI-\new{based} systems leverage advanced emotional recognition, speech analysis, and behavioral pattern detection to assess mental health conditions in real time~\cite{guemghar2022social,moulya2022mental}. Virtual therapists, driven by AI, provide personalized cognitive behavioral therapy (CBT), mindfulness training, and emotional counseling, making mental health support more accessible to underserved populations~\cite{robinson2024brief, klkeczek2024robots, omarov2023artificial, 10150665, jeong2023robotic}. Beyond digital therapy platforms, EmAI is being integrated into immersive environments, such as VR-based exposure therapy for phobias or PTSD~\cite{khatri2024virtual, gaina2024safevr, liu2022virtual, emmelkamp2021virtual, juvrik2024vret}, \new{providing} dynamic adjustments to therapy based on physiological feedback like heart rate or eye tracking. EmAI also enhances long-term monitoring through wearable technologies, detecting early warning signs of mental health decline and enabling timely interventions~\cite{alves2022robot, gedam2021review, kang2022wearable, sheikh2021wearable}, creating a seamless ecosystem of continuous, adaptive, and personalized mental health care.

\subsubsection{Post-Intervention Stage}

The field of postoperative rehabilitation is crucial for enhancing patient recovery and quality of life after surgical interventions or disease treatments. Traditional methods, while effective, often require intensive human resources and cannot always offer customized therapy. EmAI introduces a promising solution to rehabilitative care, providing continuous, adaptive, and patient-centered care~\cite{baranyi2022ai}.

\textit{\textbf{Intelligent exoskeleton.}} \new{Emerging} EmAI systems~\cite{tang2021artificial, lanotte2023ai} have been designed to assist patients in regaining mobility and strength during physical rehabilitation, offering guided exercises, precise motion assistance, and adaptive feedback for individuals with mobility impairments. For example, exoskeleton robots help patients with gait training~\cite{chang2020ai, harris2022survey}, and hand rehabilitation robots assist in restoring hand function~\cite{neog2024hand}. Upper limb rehabilitation robots have been designed to support patients’ arms during daily activities, promoting neuroplasticity and functional recovery~\cite{zhao2020new, qassim2020review, zimmermann2019anyexo}. These technologies are particularly beneficial for stroke or hemiplegia survivors and individuals with spinal cord injuries, offering them the potential for increased independence and improved quality of life~\cite{rahman2022ai}. These applications are typically developed with fine control approaches and interaction strategies~\cite{vianello2024exoskeleton, xiao2023ai}.

\textit{\textbf{Customized recovery coach.}} For balance exercises on trunk rehabilitation robots, EmAI systems~\cite{ai2022deep, lee2023machine} have shown the ability to automatically capture real-time motions and assess patients' conditions. They adjust controllers and optimize training intensity according to the individual capabilities and needs of each patient, thereby enhancing personalized rehabilitation sessions. In addition, EmAI systems with the ability to continuously learn can evolve from patients' past behavior and responses, improving their adaptability and responsiveness to patient-specific therapeutic needs~\cite{velez2021artificial, ai2021machine}. Moreover, EmAI systems can provide and update safe and personalized rehabilitation training regimens. In particular, they can predict critical rehabilitation metrics~\cite{zhang2023intelligent}, enable more accurate assessment of patient recovery, and recommend optimized postoperative rehabilitation plans~\cite{gong2023application, 9025219} based on real-time monitoring of patient motor performance and physiological feedback~\cite{swarnakar2023artificial, dias2020innovative, cha2021deep}.

\textit{\textbf{Medication controller.}} Intelligent drug delivery systems, integrated with real-time monitoring devices, adjust medication dosages dynamically based on physiological feedback such as glucose levels, blood pressure, or neural activity~\cite{raikar2023advances, wang2021emerging}. Robotic systems equipped with EmAI are being deployed for precise administration of complex therapies, such as chemotherapy or insulin delivery, minimizing errors and improving patient outcomes~\cite{shahbazi2020self,mazidi2022smart, zhang2024smart, sharma2021intelligent}. Additionally, AI-powered medication adherence tools, including smart pill dispensers and tracking apps, ensure patients follow their prescribed regimens, reducing the risks associated with missed doses or overmedication~\cite{rosdi2021smart, thangam2024smart, minera2023smart, rajan2021smart}. With predictive analytics, EmAI systems can also identify potential adverse drug interactions or recommend adjustments in real time, offering safe, efficient, and personalized pharmacological care.

\textit{\textbf{Health monitoring wearable.}} Smart wearable devices are increasingly integrated with EmAI systems, utilizing MLLMs \new{capabilities}~\cite{ho2024remoni} and enabling the real-time tracking of vital health metrics such as heart rate, oxygen saturation, and electrodermal activity~\cite{chan2009smart, lu2020wearable, pino2018wearable}. These devices can dynamically adjust to the unique physiological profiles of individual patients, offering tailored health insights and alerts~\cite{poongodi2022diagnosis}. Through continuous learning algorithms, EmAI systems evolve to better predict patient-specific health events, such as detecting early signs of infection, Parkinson’s disease, or cardiovascular issues~\cite{hafid2017full,talitckii2020avoiding}. Furthermore, these wearables can communicate with healthcare providers to ensure timely interventions, enhancing patient safety and recovery outcomes~\cite{greiwe2020wearable}. By leveraging real-time data and EmAI, these wearables not only monitor but also proactively manage postoperative care, optimizing recovery trajectories.

\textit{\textbf{Cognitive rehabilitation tool.}} In the realm of postoperative cognitive recovery, EmAI-equipped tools can customize cognitive exercises based on real-time analysis of patients' performance and progress~\cite{yuan2021systematic, greiwe2020wearable}. By integrating sensors and interactive software, EmAI systems provide a responsive platform that adjusts tasks according to cognitive load and patient capability~\cite{ryu2023study}. The continuous adaptation helps \new{create} highly effective rehabilitation sessions that can address specific cognitive deficits more precisely~\cite{almeida2020ai}. Additionally, EmAI tools can predict and monitor cognitive recovery trajectories, offering insights that guide further therapy adjustments~\cite{sale2018predicting}. Such tools not only support faster cognitive rehabilitation but also ensure it is engaging and aligned with patients' specific therapeutic needs.

\textit{\textbf{Neural prosthetics.}} Cutting-edge studies have also explored the integration of Brain-Computer Interfaces (BCIs) to enhance motor imagery (MI) training and rehabilitation. In MI training, patients imagine specific movements, and the BCI translates these intentions into actions through signal analysis such as EEG~\cite{lun2023combined} or functional brain connectivity~\cite{alanisespinosa2019assesmentfunctionalconnectivityimmersive} within a virtual environment~\cite{reyhanimasoleh2019navigatingvirtualrealityusing} or through external robotic devices~\cite{kwon2024visualmotionimageryclassification}. This method has been shown to facilitate neurological remodeling and enhance motor function in stroke survivors.

\subsection{Daily Care \& Companionship}

\subsubsection{Assistive Robot}
The integration of EmAI into healthcare has led to the development of assistive robots that enhance daily care and support \new{for} patients. 
This section focuses on three key applications, including social assistance, daily living assistance, and mobility assistance.
\vspace{7pt}

\noindent
\begin{minipage}[t]{0.44\textwidth}
    \hspace{12pt} \textit{\textbf{Social guide.}} In contemporary society, individuals with Autism Spectrum Disorder (ASD), Bipolar Disorder, or other social challenges often face significant barriers in accessing adequate social support. The advancement of EmAI systems, especially those driven by LLMs, has facilitated the development of innovative solutions aimed at addressing these needs~\cite{cooper2020ari, cooper2021social}. The NAO robot and QTrobot have been utilized to enhance facial recognition and improve eye contact, serving as therapeutic mediators for people with autism and helping them learn to interact with others~\cite{robaczewski2021socially, she2021enhance, amirova202110, el2022initial}. Through structured interactions, children can practice social skills such as imitation, turn-taking, engagement, and empathy~\cite{salhi2022towards}. 
\end{minipage}\hfill
\vspace{7pt}
\begin{minipage}[t]{0.53\textwidth}
    \vspace{-5pt}
    \centering
    \includegraphics[width=\textwidth]{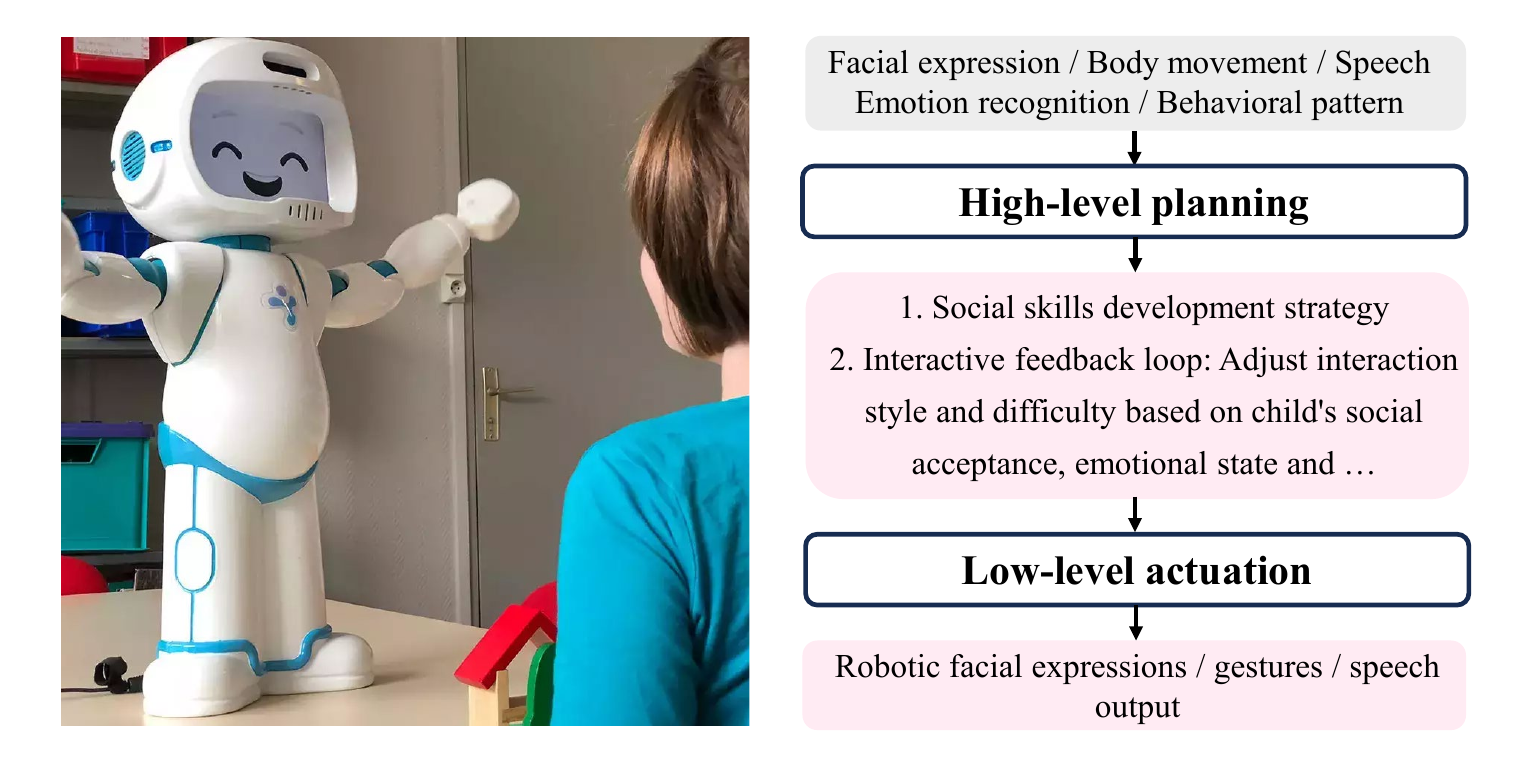}
    \captionsetup{skip=5pt} 
    \captionof{figure}{An EmAI system for socially assistive robotics. It enhances children's social skills through adaptive interaction. The system assesses and responds to children's emotional cues and social behaviors by combining high-level planning and low-level actuation.
    \textcolor{myColor}{\textbf{Source:} Reproduced from \cite{ANI2022robots}. Copyright @2025 ETHealthworld.com.}}
    \label{fig6}
    \vspace{-1em}
\end{minipage}
\vspace{5pt}

Studies have indicated that children with ASD often respond positively to robot-assisted therapy, showing increased engagement and reduced anxiety during sessions~\cite{cao2019robot, rakhymbayeva2021long, karim2023evaluate}. Beyond psychological disorders, EmAI systems like ZORA robots~\cite{van2020zora, melkas2020impacts, scerri2021formal} have \new{also helped} children with severe physical disabilities \new{achieve} therapeutic and educational goals, as well as helping patients with dementia \new{improve their} communication. We present \new{an EmAI system pipeline} that supports children with social disorders, as shown in Figure~\ref{fig6}.

\textit{\textbf{Daily helper.}} For patients facing challenges in independent living, advancements in EmAI have led to the development of various assistive robots designed to support essential daily activities. These robots assist with specific tasks such as eating, dressing, personal hygiene, and medication management. For instance, feeder robots have been developed to help severely disabled patients with self-feeding~\cite{park2020active, bhattacharjee2020more}. These systems integrate infrared sensors to achieve precise spoon control based on the user's body size and head position, automatically adjusting as needed~\cite{nawaz2021research, shih2020smart, zhi2022multimodal}. This adaptability allows users to independently select food, enhancing their dining experience and overall quality of life. Additionally, robots like ARI~\cite{cooper2020ari, cooper2021social} are designed to assist in rehabilitation programs. These robots can demonstrate exercises, offer real-time guidance, and provide encouragement, fostering greater patient engagement and adherence to therapeutic regimens~\cite{lemaignan2023open}. \new{By supporting both physical recovery and emotional well-being, these assistive technologies are becoming essential in helping patients regain independence and improve their quality of life.}

\textit{\textbf{Locomotion aide.}} EmAI has significantly advanced mobility and rehabilitation assistance for individuals with physical impairments, offering innovative solutions to restore functionality and support individuals during recovery~\cite{sahoo2023ai}. For instance, exoskeletons, such as those developed by Ekso Bionics, empower patients with spinal cord injuries to stand and walk, promoting rehabilitation and improving mobility~\cite{read2020physiotherapists}. Similarly, ReWalk~\cite{zeilig2012safety} enables individuals with paralysis to walk and climb stairs, effectively compensating for physical impairments and delivering positive outcomes. AI-powered wheelchairs demonstrate how EmAI enhances mobility. They use AI algorithms to predict movements and assist users with severe motor or cognitive impairments~\cite{sun2023survey, guo2023empowering}. These systems analyze past movements, detect obstacles, and enable safe, real-time navigation, helping users intuitively maneuver through complex environments~\cite{jameel2022wheelchair, arshad2023multi}. \new{Additionally,} EmAI-based wheelchairs are noted for their cost-effectiveness, achieved by optimizing control parameters to minimize energy consumption and extend battery life~\cite{castro2016promise}. For mobile balance-assistive robots, \new{these systems} are being developed to help individuals maintain balance and prevent falls during daily activities~\cite{wang2023gracefuluserfollowingmobile}. Some robots, such as SoloWalk~\cite{mccormick2016power}, provide weight support as patients walk, further aiding the rehabilitation process. Brain-computer interface-controlled robots, including wheelchair-mounted robotic arms~\cite{palankar2009control} and robotic prosthetics, also play a vital role in enhancing functionality for individuals with paralysis or limb loss, enabling them to perform manipulation tasks and regain independence.

\subsubsection{Companion Robot}
Companion robots powered by EmAI systems support healthcare in several key scenarios: emotional support for mental health~\cite{kokkonen2023beyond,bilalpur2024learning,ahmad2022designing,li2023systematic}, developmental support for child well-being~\cite{scassellati2012robots,pennisi2016autism,canamero2016making,huijnen2017implement,mengoni2017feasibility,grossard2018ict,cao2025workshop}, and disease monitoring support for elderly care~\cite{mesquita2016use,mesquita2017social,carrera2017dynamic,khosla2017human,chu2017service}. 
For mental health, these robots offer emotional support by engaging users in conversations, providing empathy, and fostering a sense of companionship, which can be particularly beneficial for individuals experiencing loneliness or anxiety. In children’s care, these robots engage young users in educational activities, social interaction, and play, supporting cognitive and social development in a safe, monitored environment. In elderly and chronic disease care, companion robots assist with daily activities, medication reminders, and physical monitoring, helping to improve quality of life and enabling seniors to maintain independence. They can also detect changes in health conditions, allowing for timely interventions. Through these applications, companion robots enhance mental and physical well-being, contributing meaningfully to both individual care and broader support networks.

\textit{\textbf{Emotional companion.}} Companion robots for emotional support typically operate through two main pathways: virtually EmAI agents~\cite{denecke2021artificial} and advanced AI-robotics innovations used clinically~\cite{altameem2022deep}. Virtually embodied agents, such as chatbots or conversational systems, provide accessible emotional support by engaging users in real-time, empathetic conversations that help relieve loneliness, anxiety, or stress~\cite{kokkonen2023beyond, bilalpur2024learning}. These virtual companions are easily accessible and offer immediate, low-barrier interaction for those seeking support. In clinical settings, AI-integrated physical robots take mental health support a step further by interacting with users face-to-face, recognizing physical and emotional cues, and adapting responses to offer personalized comfort, social engagement, and therapeutic activities~\cite{pham2022artificial}. Together, these virtual and physical AI advancements form a well-rounded support system, meeting both immediate conversational needs and offering in-depth emotional support in clinical environments.

Beyond virtual therapeutic applications, clinicians and researchers are working to bring AI-robotics innovations directly into clinical settings. For instance, intelligent animal-like robots, such as Paro, a plush harp seal, are increasingly used to assist patients with dementia~\cite{griffiths2014paro,yu2015use}. Alongside Paro, the larger eBear is part of a class of ``companion bots" designed to serve as at-home health aides, responding to speech and movement with interactive “dialogue” and offering companionship. These robots aim to help elderly, isolated, or depressed individuals through comforting interactions. Several studies have examined the role of such robots in reducing stress, loneliness, and agitation and in improving mood and social connections~\cite{bemelmans2012socially}.

\textit{\textbf{Kid health guardian.}}
AI-driven robots and wearable devices \new{play a key role} in diagnosing developmental disorders, monitoring vital signs, and engaging children in interactive exercises~\cite{beran2013reducing,riek2017healthcare,cao2025workshop}. By \new{incorporating AI into interactive, physical forms, these embodied systems} can adjust to a child’s responses in real-time, offering adaptive support that traditional methods may lack. In therapeutic contexts, social robots \new{help} children with autism by \new{promoting} social engagement and enhancing communication skills through controlled, repetitive interactions~\cite{scassellati2012robots}. Additionally, AI-\new{based} tools for rehabilitation and physical therapy provide targeted exercises with engaging feedback to support children with motor skill challenges~\cite{belpaeme2018social}. Overall, EmAI \new{is revolutionizing pediatric healthcare by delivering} personalized, interactive, and effective treatments that address both the physical and cognitive needs of children.

AI robots present valuable opportunities for engaging children with ASD, offering a unique approach to social skill development~\cite{grossard2018ict}. Studies show that children with autism often respond well to robots, even when they struggle to interact with others~\cite{scassellati2012robots}. The Kaspar robot, for instance, has shown potential for integration into educational and therapeutic settings~\cite{huijnen2017implement} to help improve social skills~\cite{mengoni2017feasibility}. Early studies \new{suggest that} children with ASDs often respond more positively to robot companions than human therapists, demonstrating increased social behaviors and improved spontaneous language during sessions~\cite{pennisi2016autism}. \new{Furthermore}, the development of social robots also offers a promising approach to supporting children with diabetes~\cite{canamero2016making}. \new{Such robots} not only assist with health management tasks, such as reminding users to check glucose levels and guiding insulin administration, but also provide emotional support by engaging in friendly, human-like interactions. Designed with input from children and clinicians, the robot fosters social engagement, reducing anxiety around diabetes management and improving adherence to health routines. Early findings \new{are encouraging, showing that these robots} effectively combine practical health assistance with \new{the benefits of} companionship.

\textit{\textbf{Elderly health caregiver.}} 
With aging, the elderly face numerous challenges, including declining physical health and increased vulnerability to chronic diseases. Conditions such as cardiovascular disease, diabetes, and arthritis not only diminish their quality of life but also impose considerable stress on healthcare systems and caregivers. These persistent health issues frequently result in reduced mobility, social isolation, and a dependence on others for daily activities, further exacerbating their vulnerability. In light of these challenges, it is crucial to explore innovative solutions. EmAI-based elderly care robots, with advanced long-term learning capabilities, can adapt to an individual's health trajectory, habits, and preferences over time. This long-term learning capability may \new{include integrating} RL for personalized care optimization or meta-learning for rapid adaptation, as well as the application of disease-specific predictive modeling to address the evolving nature of elderly users' needs.

EmAI has been shown to offer practical and emotional support to enhance the outcomes of elderly patients. Robotics with EmAI capabilities assist elderly individuals by improving mobility, conducting physiological assessments, and monitoring vital signs, which \new{helps with} daily activities, rehabilitation exercises, and independent navigation~\cite{riek2017healthcare}. Studies show that elderly patients feel more secure and comfortable with companion robots, as they can be continuously monitored without intruding on their personal space or autonomy~\cite{niemela2021towards}. By combining EmAI monitoring systems, elderly companion robots provide a reliable, non-intrusive way to support patient health, making them a valuable asset in home-based care.

Additionally, these AI-enabled robots enhance the quality of care for dementia patients, providing cognitive engagement and symptom management through structured activities and consistent interactions that traditional care methods may not \new{be able to} offer. The majority of elderly individuals with dementia who engaged with social robots Sophie and Jack experienced highly positive interactions~\cite{chu2017service}. These robots are equipped with human-like communication abilities, emotional expressions, and gestures, and they can play songs, games, and tell stories. They have been shown to provide sensory enrichment and foster social engagement for dementia patients. For instance, many participants were motivated to play bingo with Jack and actively joined other group activities, highlighting the robots' impact on promoting social and cognitive engagement~\cite{khosla2017human}. 

Another key role of embodied AI is to provide companionship, alleviating feelings of loneliness and fostering emotional well-being among elderly patients, which is crucial for mental health and quality of life~\cite{kachouie2014socially}. Studies~\cite{robinson2013psychosocial,robinson2013suitability,robinson2014role} have shown that participants who interacted with robots Paro or AIBO experienced a significant reduction in their systolic and diastolic blood pressure, and loneliness~\cite{banks2008animal} over the trial period, in contrast to those in the control group. Overall, EmAI is emerging as a vital tool in elderly care, addressing complex physical, cognitive, and emotional needs.

\subsection{Infrastructure Support}
\subsubsection{Rescue Robot}
\vspace{7pt}

\noindent
\begin{minipage}[t]{0.44\textwidth}
    \hspace{12pt} EmAI-\new{based} robots are increasingly \new{being} developed to serve as human assistants or replacements in various high-risk environments. In rescue scenarios, their potential has been widely explored due to their ability to operate safely in dangerous environments and urgent conditions. In the aftermath of natural disasters or life-threatening emergencies, search and rescue robots aim to provide rapid aid and emergency health services to victims. Thanks to the advancement of recent AI approaches including MLLMs and VLMs, EmAI-equipped robots can efficiently locate and assist survivors while supporting rescue operations~\cite{fung2024mllm, shah2023lm, mahmud2024atr, shree2021exploiting} in disaster-prone areas, such as earthquake zones. A representative workflow of EmAI-based rescue robots is illustrated in Figure~\ref{fig7}. 
\end{minipage}\hfill
\vspace{7pt}
\begin{minipage}[t]{0.53\textwidth}
    \vspace{-5pt}
    \centering
    \includegraphics[width=\textwidth]{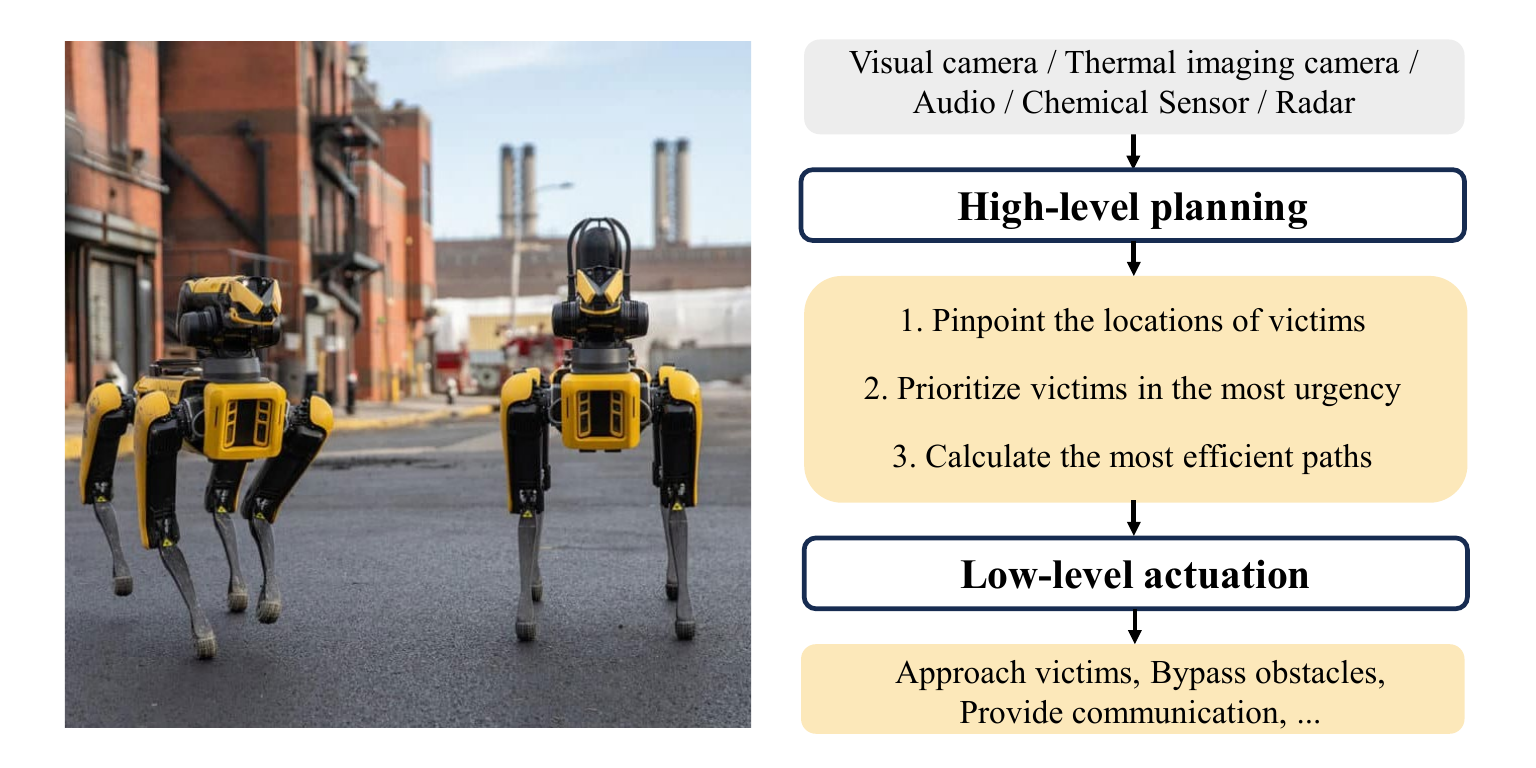}
    \captionsetup{skip=5pt} 
    \captionof{figure}{An EmAI system in rescue robotics: optimizing search and rescue operations. This system integrates sensory data with strategic processing to quickly identify and prioritize victims, effectively navigating through complex environments to facilitate timely and efficient rescue missions. \textcolor{myColor}{\textbf{Source:} Reproduced from \cite{BostonDynamics2024SpotRescue}. Copyright ©2024 Boston Dynamics.}}
    \label{fig7}
    \vspace{-1em}
\end{minipage}
\vspace{5pt}

\textit{\textbf{Life scout.}} Technological advances have \new{produced} sophisticated survivor detection systems, which can quickly identify and locate survivors in emergencies using integrated sensors and drones~\cite{dong2021uav, cardona2021visual}. These systems leverage thermal and RGB imaging, \new{along with AI for real-time data analysis}, efficiently pinpointing \new{survivors}~\cite{cruz2021autonomous, wang2023development}. Recent developments in AI and robotics have enabled systems to analyze vast amounts of data quickly, making real-time survivor detection more accurate and reliable~\cite{sambolek2021automatic}. By integrating \new{both} visual and audio signals through a wireless sensor network, established EmAI systems~\cite{sharma2022real, cruz2021autonomous} are able to detect survivors’ locations, transmit their position \new{information} to a centralized cloud server, and assess environmental risk levels to \new{support rescue planning}.

\textit{\textbf{Agile walker.}} The life-search ability of rescue robots has been enhanced by AI-based mobility systems that are capable of operating in both hazardous and confined environments. In such environments, high mobility is essential for navigating rough terrain, clearing obstacles, and streaming live video from affected areas~\cite{tong2024robots, li2024survey}. Current advancements aim to enable autonomous navigation by equipping these robots with AI-based obstacle detection and avoidance capabilities~\cite{bravo2021internet, papyan2024ai, ismail2021military}. 
Legged robots, particularly quadrupeds equipped with advanced EmAI systems, have demonstrated the ability to navigate rough surfaces, debris, and areas with variable friction, making them highly adaptable in complex terrains~\cite{smith2022legged}.  \new{Furthermore}, recent RL-augmented frameworks further enhance robots in adaptive blind locomotion, high-speed movement, dynamic response to obstacles, and adaptability improvement under uncertain conditions~\cite{chen2024learning,bindu20203}. 

\textit{\textbf{Rescue carrier.}} At the current stage, most rescue robots remain limited to data retrieval and collection tasks~\cite{bindu2019sigma, hong2018development, ulloa2022design, edlinger2022innovative}. In the next phase, it is crucial to integrate multi-degree-of-freedom manipulators based on advanced EmAI systems, such as VLA models, to further expand the functionality of rescue robots. \new{These enhancements would enable them to hold, grasp, and transfer objects}, as well as \new{assist} victims with medical equipment. With these medical support tools, rescue robots \new{could} offer direct on-site assistance to victims~\cite{cruz2023mixed}. 

\subsubsection{Delivery Robot.}
Delivery robots are now widely used in healthcare settings, performing in-hospital delivery tasks that were previously handled by humans, thus overcoming limitations related to time constraints and availability. With EmAI systems, they are capable of delivering essential items—such as samples, meals, medications, and medical supplies~\cite{patel2021evolution, maity2022review, srinivas2022autonomous} more efficiently, supporting healthcare facilities in streamlining operations and alleviating staff burdens~\cite{hossain2023autonomous}. These robots can independently navigate complex environments, adapt to dynamic conditions, and carry out functions like obstacle avoidance~\cite{law2021case}, speech interaction~\cite{grasse2021speech}, and precise patient facial recognition~\cite{geethanjali2024ai}. During the COVID-19 pandemic, robots have also been deployed for tasks like the delivery of medications and food~\cite{patel2021medbuddy, sarker2021robotics}, thereby minimizing human contact and \new{reducing} the risk of virus transmission.

\subsubsection{Disinfection Robot.}
The pandemic has also significantly increased the demand for service robots to replace human labor in contaminated areas, particularly for disinfection tasks. This has made efficient and rapid disinfection in hospitals crucial, sparking research into disinfection robots. By integrating mobile robot platforms with advanced disinfection technologies, such as hydrogen peroxide atomization~\cite{ruan2021smart} and UVC devices~\cite{guettari2021uvc, mehta2023uv}, and equipping them with EmAI navigation systems, these robots are able to automate cleaning in complex indoor environments. They are designed to disinfect both floor surfaces and air by eliminating harmful pathogens using a combination of vacuum cleaners, sanitizers, and UV light disinfection methods~\cite{kim2019control, rai2020autonomous, nosirov2020design, zaman2022uvc}. By upholding rigorous hygiene standards, these robots significantly mitigate the risk of cross-contamination in healthcare settings.
\vspace{17pt}

\noindent
\begin{minipage}[t]{0.44\textwidth}
    \subsection{Biomedical Research}
    \vspace{5pt}
    \hspace{12pt} EmAI also shows great potential in biomedical and healthcare research~\cite{wang2023scientific}. It can synergistically combine human creativity and expertise with AI’s capabilities in handling vast biomedical datasets, navigating complex hypothesis spaces, and executing repetitive experiments. The core functionality of EmAI systems in biomedical research involves planning discovery workflows, performing self-assessments to recognize and address knowledge gaps, and employing structured memory for continual learning~\cite{gao2024empowering, wang2023accelerating}. This workflow is illustrated in Figure~\ref{fig8}, which shows an automation process of chemical experiments.
\end{minipage}\hfill
\vspace{7pt}
\begin{minipage}[t]{0.53\textwidth}
    \vspace{-7pt}
    \centering
    \includegraphics[width=\textwidth]{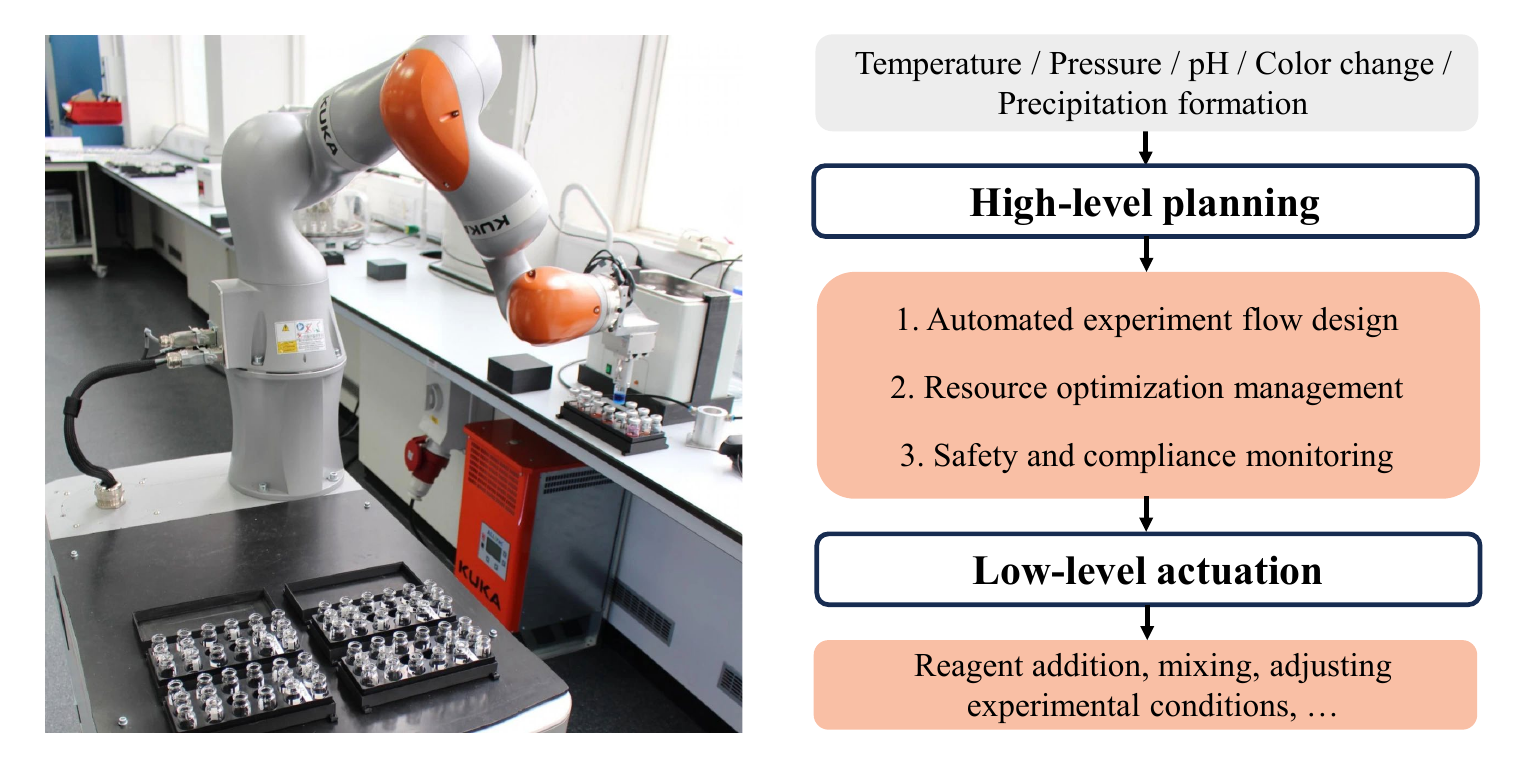}
    \captionsetup{skip=5pt} 
    \captionof{figure}{An EmAI system in laboratory automation: enhancing experimental precision and efficiency. This setup utilizes advanced planning algorithms for experiment design and resource management, paired with robotic actuators for precise manipulation of reagents and experimental conditions, ensuring compliance with safety standards.
    \textcolor{myColor}{\textbf{Source:} Reproduced from~\cite{burger2020mobile} with permission from Springer
Nature.}}
    \label{fig8}
    \vspace{-1em}
\end{minipage}
\vspace{7pt}

\textit{\textbf{Automated lab technician.}} Experiments are critical \new{components} in biomedical research. EmAI-driven robots automate repetitive tasks, increasing throughput and allowing researchers to focus on more complex analysis tasks. Scientists have created robot chemists capable of autonomously conducting complex chemical reactions and analyses~\cite{burger2020mobile, song2024multi}. These robotic systems apply EmAI in laboratory automation, enabling them to design experiments~\cite{Zhu2022An, Szymanski2021Toward, Coley2019A}, execute procedures~\cite{Burger2020A, Ha2023AI-driven}, and interpret results~\cite{Kalinin2021Automated, Groth2017Indicators, Wang2018Light-Driven} without human intervention. For instance, a platform leverages AI-based synthesis planning, trained on millions of reactions, to propose synthetic routes, which are then fine-tuned by expert chemists and executed on a robotic flow system~\cite{coley2019robotic}. This setup allows for scalable, reproducible synthesis of a variety of compounds, such as pharmaceuticals. Another AI-driven robotic chemist also demonstrated its ability to autonomously perform tasks ranging from synthetic planning to executing experiments in batch reactors~\cite{ha2023ai}. In the field of materials science for photo-catalytic hydrogen production, the robot operates autonomously in a laboratory environment, using Bayesian optimization to navigate a complex experimental space with ten variables, performing 688 experiments over eight days~\cite{burger2020mobile}. In summary, EmAI can combine precision handling with autonomous decision-making, allowing it to run experiments continuously without human intervention. These experiment setups minimize human bias, accelerate discovery and production, and enable exploration of large experimental spaces.

\textit{\textbf{Pioneer in drug discovery.}} The traditional drug discovery process is typically time-consuming and involves substantial costs, presenting significant challenges for pharmaceutical research and development. However, the integration of EmAI into the drug discovery process not only shortens the overall development timeline but also reduces the associated expenses. AI-driven robotic systems can perform rapid screening of vast chemical libraries, identifying potential drug candidates more efficiently than manual methods~\cite{mak2024artificial}. In high-throughput screening (HTS) systems, physical components are primarily responsible for automating processes such as sample handling, reaction setup, and data collection through robotics and specialized software for instrument control~\cite{michael2008robotic}. By employing liquid handling devices, robotic technology, detectors, and specialized software, researchers have also achieved automation and high efficiency in HTS processes~\cite{xiao2021high} to swiftly evaluate large libraries of compounds for activity against specific biological targets. Additionally, in the analysis and prediction of biological sequences, such as DNA, RNA, or protein structures, LLMs and other approaches are applied to decipher patterns in biological sequences~\cite{zhou2024protclip}. By treating nucleotides and amino acids as words, they can predict biological structures and functions, thus aiding in the discovery of disease biomarkers and drug targets~\cite{wang2023pre, gu2021domain, lee2020biobert}. Similarly, AI models have revolutionized drug discovery by enhancing various processes, including toxicity prediction, drug release monitoring, Quantitative Structure-Activity Relationship (QSAR) analysis, drug repositioning, physiochemical property prediction, \etc~\cite{parvatikar2023artificial}. By leveraging RL algorithms, AI-driven drug discovery enhances both virtual screening and de novo drug design. Virtual screening~\cite{carpenter2018machine, zhang2022design, gao2024drugclip} uses RL to identify potential drug candidates from existing compound databases, while de novo design~\cite{wang2022deep, liu2021drugex, atance2022novo, korshunova2022generative} generates novel molecules with optimized biological activity and properties. By integrating RL with active learning and predictive modeling, these systems enhance exploration in vast chemical spaces, reducing experimental costs and improving hit-to-lead timelines~\cite{elbadawi2021advanced, yang2021hit}. Although they are not fully EmAI systems, they offer valuable insights and core components for the development of future EmAI-driven drug discovery platforms and possess the potential to integrate with physical devices.

\textit{\textbf{AI-based knowledge retriever.}} As EmAI continues to evolve, \new{an increasingly important focus} is \new{on developing} systems that \new{mimic} the analytical capabilities of human scientists. Advanced frameworks like ChemCrow~\cite{m2024augmenting} and CALMS~\cite{prince2024opportunities} exemplify this trend, utilizing LLMs such as GPT-4, multimodal generative models, and retrieval approaches like GraphRAG~\cite{edge2024local} to integrate scientific knowledge, biological principles, and theoretical frameworks~\cite{sanderson2023gpt, tian2024opportunities, chakraborty2025prompt}. These systems go beyond traditional data processing. They engage in critical thinking by questioning assumptions, evaluating evidence, and validating conclusions. In addition, researchers \new{have} integrated large amounts of data from publications in PubMed~\cite{zhang2023biomedclip,luo2022biogpt}, Wikipedia~\cite{guo2020wiki} and other open resources to pretrain generative models. These generative models can function as part of a composite system, integrating EmAI world models, experimental platforms, and human expertise to tackle complex problems that require interdisciplinary approaches~\cite{gao2024empowering}. In summary, the development of AI knowledge retriever marks a promising future direction in which EmAI could play an even more integral role in advancing knowledge and innovation across diverse fields.

\subsection{Summary}
We outlined the applications of EmAI in healthcare across four \new{key} domains: clinical intervention, daily care \& companionship, infrastructure support, and biomedical research. These domains are inherently interdependent, collectively forming the cornerstone of human healthcare needs, and EmAI begins to play increasingly vital roles in some of them. As advancements in medical technology and life sciences progress, the boundaries between these domains are becoming \new{more} porous. An EmAI system for daily care \& companionship may act as an extension of an EmAI system for clinical intervention, and an EmAI with monitoring information for daily care \& companionship can offer valuable pre-intervention data for clinical intervention.

In the early stages of EmAI development, limitations in data acquisition and capabilities of multimodal large-scale models, as well as the uneven maturation of EmAI-related research subfields, led to fragmented applications (as we introduced in this section), where an EmAI system could only benefit a specific scenario or sub-scenario. For instance, certain EmAI systems might have been capable of performing image analysis  \new{yet could not plan} surgical procedures based on the analysis. However, they share similar backend AI approaches such as world models, LLMs and MLLMs. As EmAI continues to evolve, there is promising potential for these applications to merge,  \new{paving the way for more holistic and versatile EmAI systems that} meet the comprehensive needs across all four healthcare domains.

\begin{figure}[ht!]
    \centering
    \includegraphics[width=\linewidth]{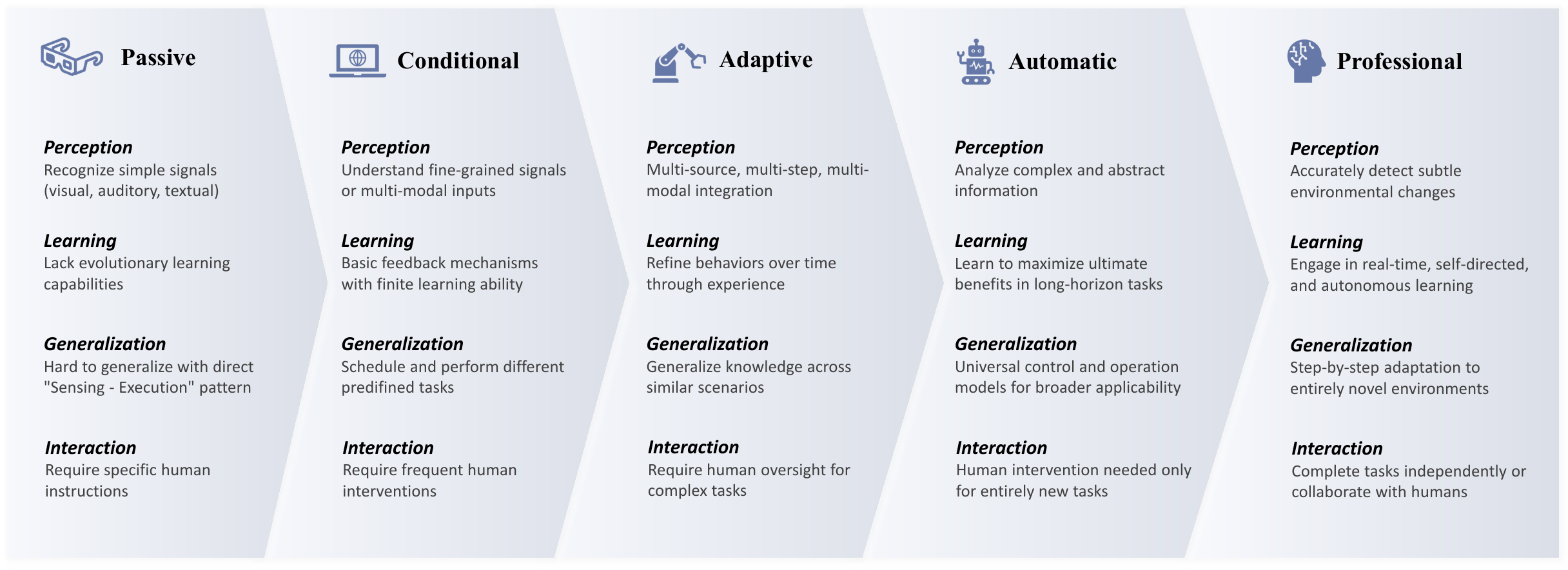}
    \caption{Levels of embodied AI. We outline the capabilities of embodied AI across different levels, viewed from four perspectives: perception, learning, task generalization (or generalization), and human-machine interaction (or interaction).}
    \label{fig9}
\end{figure}

\section{Intelligent Levels of Embodied AI}
\label{section4}

Clearly defining the levels of embodied AI provides explicit guidance for the development of embodied AI frameworks and application-oriented products. The proposed reference levels delineate the stages of evolution and articulate the capabilities of embodied AI as they progress through Levels 1 to 5, which emphasize incremental advancements, spanning 4 aspects of perception, learning capabilities, task generalization, and human-machine interaction, as summarized in Fig~\ref{fig9}.

\noindent \textbf{Level 1: Passive} - Capable of receiving and recognizing simple signals (visual, auditory, textual, and tactile) and executing limited, straightforward actions, including basic movements and responses, while lacking evolutionary learning capabilities. Demonstrate a direct ``\textit{Sensing - Execution}'' pattern that is difficult to generalize to other tasks, with a restricted connection between perception and action. Nearly all tasks require specific instructions, interventions, or collaborations from humans.

\noindent \textbf{Level 2: Conditional} - Exhibit enhanced perceptual capabilities for understanding and detecting fine-grained signals (such as distance and depth) or processing multimodal inputs. Implement basic feedback mechanisms and limited learning abilities triggered by specific conditions (e.g., a robot slowing down when it detects an obstacle, or adjusting its path when the GPS signal is weak), with a restricted capacity to interact with the environment. Capable of scheduling and performing predefined tasks, but human intervention is frequently required to establish feedback mechanisms or assist in completing complex tasks.

\noindent \textbf{Level 3: Adaptive} - Demonstrate improved multi-source, multi-step, multi-modal perceptual integration and decision-making capabilities. Capable of continuously adapting and refining behaviors over time through experience based on interactions with the environment and evolutionary learning mechanisms. This level enables the system to generalize knowledge across similar scenarios, starting with varied operations on the same object, progressing to identical operations on different objects, and eventually handling diverse operations on various objects. However, human supervision remains necessary for complex tasks or novel environments.

\noindent \textbf{Level 4: Automatic} - Exhibit advanced signal processing capabilities to analyze complex and abstract information across various modalities, sources, and time, while interacting with previous perceptions or knowledge stored in memory. The system can autonomously manage multiple tasks simultaneously and learn from past actions to optimize execution strategies to maximize ultimate benefits in long-horizon tasks. It introduces universal control and operation models, broadening its applicability. Human intervention is only required for unexpected challenges or entirely new tasks.

\noindent \textbf{Level 5: Professional} - Capable of precise perception of subtle environmental changes, as well as understanding high-level abstract semantic information, such as emotions. Exhibit a high degree of autonomy and flexibility, able to transfer and adapt behaviors across various tasks and domains while performing complex tasks without human supervision. Engage in real-time, self-directed, and autonomous learning, adapting to entirely novel tasks and environments step by step, without human guidance. Capable of completing tasks independently or collaborating with humans. Achieve full integration with various tools, enabling comprehensive task execution and enhancing outcomes in a way like a highly skilled human expert.

The proposed \textit{levels of embodied AI} framework offers clear definitions to evaluate and guide the development of embodied AI systems. However, an embodied AI system does not necessarily fit neatly into a single level; it may demonstrate varying degrees of capability across the four dimensions, \textit{i.e.} perception, learning, generalization, and human-machine interaction. This framework not only reflects technological advancements but also emphasizes the potential for intelligent systems to evolve and integrate seamlessly into various aspects of human life. From a practical perspective, while higher levels of intelligence often yield substantial benefits, pursuing advanced levels indiscriminately is not recommended. Instead, it is essential to strike a balance between cost and application value.

Table~\ref{examples} presents examples of EmAI systems to clearly illustrate the progression and levels of EmAI, ranging from basic operations (Level 1) to advanced autonomy and decision-making (Level 5). At present, most frameworks function at Levels 1 to 3 or focus solely on a single sub-functional module. Surgical robots, for instance, execute pre-programmed motions (Level 1), monitor vital signs and alert clinicians to abnormalities (Level 2), and integrate multimodal inputs to perform precise tasks such as suturing or injecting (Level 3). However, they have yet to achieve the autonomy of Level 4 and 5 systems, which require real-time decision-making and the ability to detect subtle anatomical variations. Similarly, companion robots provide simple auditory or touch-based responses (Level 1), recognize gestures and adapt behaviors (Level 2), and assess physical and mental health for personalized support (Level 3). Still, they fall short of understanding complex emotional states or offering proactive, personalized care at Levels 4 and 5. 
While significant progress has been made, further advancements are needed to overcome current limitations, enabling these systems to reach Levels 4 and 5, where they can perform independent reasoning, complex decision-making, and truly autonomous operations.

\section{Datasets and Benchmarks}\label{section5}

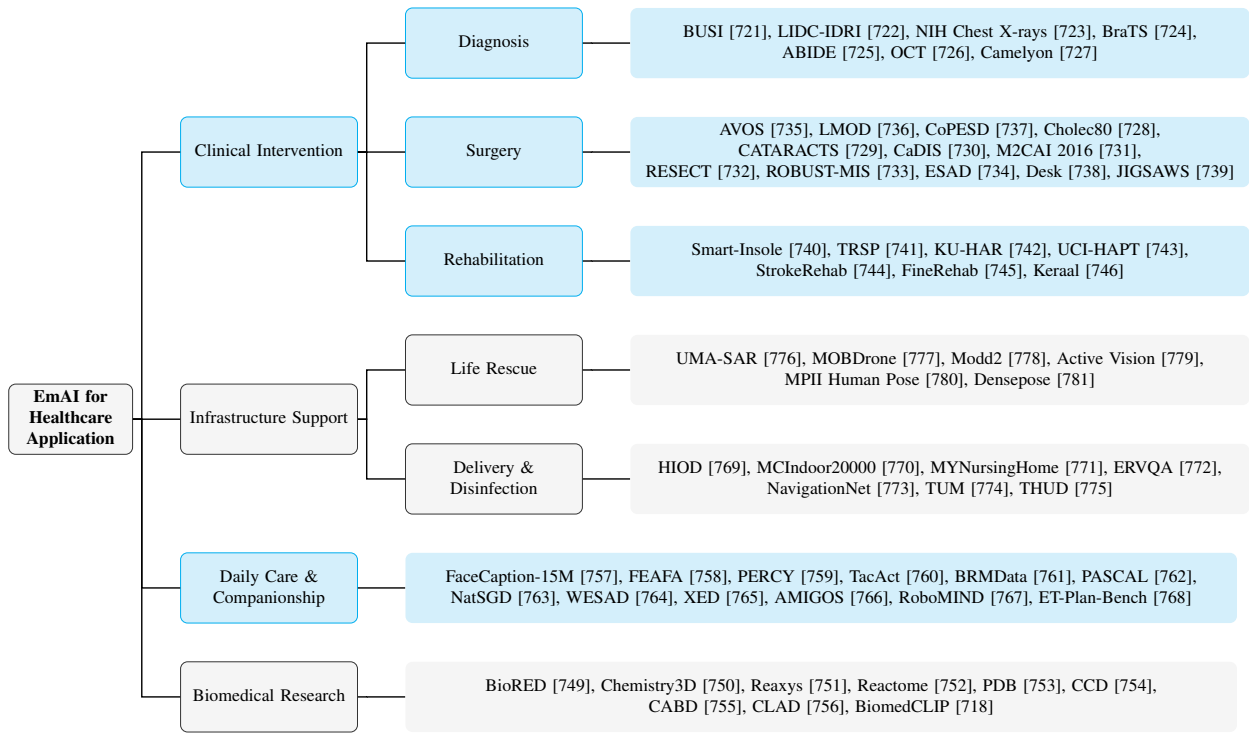
\begin{figure}[t] 
    \centering
    \resizebox{\linewidth}{!}{ 
    \begin{forest}
    for tree={
        grow=east,
        draw,
        rounded corners,
        minimum height=4em, 
        text width=35em, 
        align=center, 
        inner sep=0pt, 
        text centered, 
        parent anchor=east,
        child anchor=west,
        edge path={
            \noexpand\path [draw, thick, >=stealth]
            (!u.parent anchor) -- +(5pt,0) |- (.child anchor)\forestoption{edge label};
        },
        l sep+=15pt, 
        s sep+=15pt  
    }   
    [EmAI for\\ Healthcare\\ Application,
    font=\bfseries,
    fill=gray!30,
    draw=black!80,
    xshift=0, yshift=1.5em,
    minimum height=4em,
    text width=7em,
    align=center, 
        [Biomedical Research,
            draw=black!80,
            fill=gray!30,
            text width=10em, 
            align=center, 
            [{
            BioRED~\cite{luo2022biored},
            Chemistry3D~\cite{li2024chemistry3d},
            Reaxys~\cite{reaxys},
            Reactome~\cite{croft2010reactome},
            PDB~\cite{berman2000protein},
            CCD~\cite{westbrook2015chemical},\\
            CABD~\cite{zou2024benchmark},
            CLAD~\cite{ding2021new},
            BiomedCLIP~\cite{zhang2023biomedclip}
            },
                fill=gray!30,
                text width=47em,
                draw=none,
                align=center,
                xshift=0, yshift=0.6em,
            ]    
        ]
        [Daily Care \&\\ Companionship,
            fill=cyan!15,
            draw=cyan,
            text width=10em, 
            align=center, 
            [{FaceCaption-15M~\cite{dai202415m},
            FEAFA~\cite{yan2019feafa},
            PERCY~\cite{althubyani2024percy},
            TacAct~\cite{wang2021organization},
            BRMData~\cite{zhang2024empowering}, 
            PASCAL~\cite{gomes2013classifying},\\
            NatSGD~\cite{shrestha2024natsgd},
            WESAD~\cite{schmidt2018introducing},
            XED~\cite{ohman2020xed},
            AMIGOS~\cite{santamaria2018using},
            RoboMIND~\cite{wu2024robomind},
            ET-Plan-Bench~\cite{zhang2024plan}
            },
            fill=cyan!15,
            text width=47em,
            draw=none,
            ]
        ]
        [Infrastructure Support,
            draw=black!80,
            fill=gray!30,
            text width=10em, 
            align=center, 
            [Delivery \& \\Disinfection,
            draw=black!80,
            text width=10em,
            align=center, 
            fill=gray!30,
                [{HIOD~\cite{hu2023object}, MCIndoor20000~\cite{bashiri2018mcindoor20000}, MYNursingHome~\cite{ismail2020mynursinghome}, 
                ERVQA~\cite{ray2024ervqa},\\
                NavigationNet~\cite{huang2018navigationnet},
                TUM~\cite{sturm2012benchmark},
                THUD~\cite{tang2024mobile}
                },
                fill=gray!30,
                draw=none,
                ]
            ]
            [Life Rescue,
            text width=10em,
            draw=black!80,
            align=center, 
            fill=gray!30,
                [{UMA-SAR~\cite{morales2021sar}, MOBDrone~\cite{cafarelli2022mobdrone}, Modd2~\cite{bovcon2018stereo}, Active Vision~\cite{ammirato2017dataset},\\
                MPII Human Pose~\cite{pishchulin2014fine},
                Densepose~\cite{guler2018densepose}
                },
                fill=gray!30,
                draw=none,
                xshift=0, yshift=0.6em,
                ]
            ]
        ]
        [Clinical Intervention,
            text width=10em, 
            draw=cyan,
            fill=cyan!15,
            align=center, 
            [Rehabilitation,
            text width=10em,
            fill=cyan!15,
            draw=cyan,
            align=center, 
                [
                {Smart-Insole~\cite{chatzaki2021smart},
                TRSP~\cite{dolatabadi2017toronto},
                KU-HAR~\cite{sikder2021ku}, UCI-HAPT~\cite{reyes2014human},\\ StrokeRehab~\cite{kaku2022strokerehab},
                FineRehab~\cite{li2024finerehab},   
                Keraal~\cite{nguyen2024medical}
                },
                fill=cyan!15,
                draw=none,
                xshift=0, yshift=0.6em,
                ]
            ]
            [Surgery,
            text width=10em,
            align=center, 
            fill=cyan!15,
            draw=cyan!,
                [{AVOS~\cite{goodman2024analyzing}, LMOD~\cite{qin2024lmod}, CoPESD~\cite{wang2024copesd}, Cholec80~\cite{twinanda2016endonet},\\
                CATARACTS~\cite{al2019cataracts},
                CaDIS~\cite{grammatikopoulou2021cadis},
                M2CAI 2016~\cite{stauder2017tum},\\
                RESECT~\cite{xiao2017re},
                ROBUST-MIS~\cite{ross2021comparative},
                ESAD~\cite{bawa2021saras},
                Desk~\cite{madapana2019desk},
                JIGSAWS~\cite{ahmidi2017dataset}
                },
                fill=cyan!15,
                draw=none,
                xshift=0, yshift=1.2em,
                ]
            ]
            [Diagnosis,
            text width=10em,
            align=center, 
            fill=cyan!15,
            draw=cyan,
                [{BUSI~\cite{al2020dataset}, LIDC-IDRI~\cite{armato2011lung}, NIH Chest X-rays~\cite{national2017nih}, BraTS~\cite{baid2021rsna}, \\ABIDE~\cite{di2014autism},
                OCT~\cite{kermany2018identifying},
                Camelyon~\cite{litjens20181399}},
                fill=cyan!15,
                draw=none,
                xshift=0, yshift=0.6em,
                ]
            ]
        ]
    ]
    \end{forest}
    }
    \caption{The taxonomy of healthcare datasets and benchmarks for EmAI applications.}
    \label{fig10}
\end{figure}

High-quality datasets and benchmarks are \new{essential} for driving advancements in EmAI research within healthcare. In this section, we highlight representative datasets that are currently used—or hold promise—for training and evaluating EmAI systems in healthcare applications, as illustrated in Figure~\ref{fig10}. These datasets span a wide range of application domains, from general clinical procedures to specialized healthcare services, as detailed in Section~\ref{section3}. 

For \textit{\textbf{preoperative diagnosis}}, EmAI models are often trained on diverse medical imaging datasets~\cite{al2020dataset, armato2011lung, national2017nih, baid2021rsna, di2014autism, kermany2018identifying, litjens20181399}, \new{which enable} disease recognition and diagnostic support. In \textit{\textbf{intraoperative procedures}}, datasets comprising surgical videos or image demonstrations~\cite{twinanda2016endonet, al2019cataracts, grammatikopoulou2021cadis, stauder2017tum, xiao2017re, ross2021comparative, bawa2021saras} and VLM datasets~\cite{goodman2024analyzing, qin2024lmod, wang2024copesd} facilitate multi-modal policy learning and enhance surgical workflows. Additionally, robotic activity and kinematic datasets~\cite{madapana2019desk, ahmidi2017dataset} contribute to advancing surgical automation. As for \textit{\textbf{postoperative rehabilitation}}, human activity recognition datasets collected via wearable sensors~\cite{chatzaki2021smart, dolatabadi2017toronto, sikder2021ku, reyes2014human} enable \new{automated} healthcare monitoring. Multimodal datasets~\cite{kaku2022strokerehab, li2024finerehab, nguyen2024medical} that integrate vision and sensory inputs empower EmAI systems to analyze patient movements, supporting personalized rehabilitation planning and improving patient outcomes. A critical consideration in curating these datasets is the meticulous collection and de-identification of patient-related data to uphold ethical standards and safeguard privacy~\cite{price2019privacy, saheb2021mapping}.

\begin{landscape}
\thispagestyle{empty}
\begin{table}[htbp!] 
\scriptsize
\centering
\captionsetup{font=Large,skip=5pt}
\caption{Healthcare Examples with Key Features of Embodied AI in Different Levels.}
\label{examples}
\resizebox{1.05\linewidth}{!}{ 
\begin{tabular}{@{}p{1cm}p{3.5cm}p{3.5cm}p{3.5cm}p{3.5cm}@{}}
\toprule
\textbf{Level}  & \textbf{Surgical EmAI} & \textbf{Companion EmAI} & \textbf{Rescue EmAI} & \textbf{Research EmAI} \\ \midrule
\rowcolor{LightSkyBlue!10} Level 1 & \textbf{1)} Receive direct commands from surgeons, such as movement initiation or cessation. \textbf{2)} Execute preprogrammed motions. \textbf{3)} Operate solely under the direct control of human surgeons. & \textbf{1)} Detect basic auditory cues or touch inputs, such as clapping or tapping. \textbf{2)} Provide fixed and straightforward responses like playing a sound or moving when touched. & \textbf{1)} Detect stimuli such as light or large obstacles. \textbf{2)} Wait and follow direct human commands. \textbf{3)} Deliver medications in a single, stable environment and through predefined routes. & \textbf{1)} Collect basic data such as temperature or pH readings. \textbf{2)} Repeat predefined protocols. \textbf{3)} Requires initial human programming. 
\\ \midrule
\rowcolor{LightSkyBlue!20} Level 2 & \textbf{1)} Monitor multimodal patient vital signs like heart rate and oxygen levels. \textbf{2)} Adjust treatment if vital signs deviate. \textbf{3)} Provide alerts to clinicians. & \textbf{1)} Recognize multiple human gestures, speech,~\etc. \textbf{2)} Adapt responses in predefined behavioral patterns based on user requirements.  \textbf{3)} Provide condition-based reminders like taking medication, avoiding sedentary behavior,~\etc. & \textbf{1)} Sense obstacles through distance and depth sensors. \textbf{2)} Automatically adjust paths when encountering obstacles. \textbf{3)} Navigate within known environments. & \textbf{1)} Track and adjust factors such as pH or humidity to optimize experiments such as cell culture growth. \textbf{2)} Trigger predefined responses when conditions meet certain criteria. 
\\ \midrule
\rowcolor{LightSkyBlue!30} Level 3 & \textbf{1)} Integrate multimodal inputs from cameras and sensors. \textbf{2)} Adjust force, gesture, and motion based on tissue structure in real time. \textbf{3)} Master only one surgical skill, such as suturing or injecting. & \textbf{1)} Automatically and continuously assess user conditions including physical and mental health. \textbf{2)} Offer personalized wellness programs. \textbf{3)} Optimize responses through interactions and apply learned behaviors to similar routines. & \textbf{1)} Utilize radar, vision systems, and environmental sensors. \textbf{2)} Optimize lifesaving routes and processes. \textbf{3)} Coordinate with other robots and staff. & \textbf{1)} Simulate and predict outcomes for new experiments and assist researchers in experimental design and hypothesis testing. \textbf{2)} Analyze based on multiple sources of data and learn from experimental results to optimize protocols and improve experimental manipulations. 
\\ \midrule
\rowcolor{LightSkyBlue!40} Level 4 & \textbf{1)} Accurately detect lesion area and control robotic arm for delicate surgery. \textbf{2)} Master a wide range of surgical skills and can perform complete surgery with minimal human set-up. & \textbf{1)} Interpret complex environmental and biometric data. \textbf{2)} Learn from daily routines to anticipate needs. \textbf{3)} Proactively optimize response strategies based on long-term trends in mood and health status. & 1) Detect human presence and assess environmental conditions. 2) Monitor health indicators to identify patients in need of first aid. 3) Navigate unfamiliar and hazardous environments. & \textbf{1)} Integrate data from multiple ongoing experiments. \textbf{2)} Modify experimental parameters autonomously and in real time to optimize results. \textbf{3)} Apply successful methodologies and workflows to new research areas. \\ \midrule
\rowcolor{LightSkyBlue!50} Level 5 & \textbf{1)} Detect subtle anatomical variations and physiological responses. \textbf{2)} Improve surgical techniques through both real time perceptions and cumulative experiences. \textbf{3)} Make autonomous decisions during surgery akin to expert surgeons. & \textbf{1)} Understand complex emotional states and detect subtle changes. \textbf{2)} Develop personalized support strategies based on individual interactions. \textbf{3)} Proactively offer help like a true friend and switch support freely in different situations. & \textbf{1)} Recognize faint signs of life and risks, ensuring early warning. \textbf{2)} Master a variety of rescue operations, flexibly managing a variety of unexpected and unforeseen challenges. \textbf{3)} Fully integrate with emergency services to provide optimal response. & \textbf{1)} Interpret multiomics data and proactively search for relevant research papers and clinical trial results, forming a comprehensive knowledge base. \textbf{2)} Gain innovative scientific insights by automating experimental design, execution and data analysis. \\ \bottomrule
\end{tabular}
}
\end{table}
\end{landscape}  

For \textit{\textbf{infrastructure support}}, various datasets~\cite{morales2021sar, cafarelli2022mobdrone, bovcon2018stereo, ammirato2017dataset, pishchulin2014fine, guler2018densepose} have been developed to support tasks such as autonomous obstacle avoidance and life detection in emergency scenarios. In hospital settings, EmAI-driven medication delivery and disinfection systems are predominantly trained on indoor datasets~\cite{hu2023object, bashiri2018mcindoor20000, ismail2020mynursinghome, ray2024ervqa, huang2018navigationnet, sturm2012benchmark, tang2024mobile}, which enhance capabilities in environmental understanding, object manipulation, and task-oriented navigation. Datasets for \textit{\textbf{Daily Care \& Companionship}}~\cite{dai202415m, yan2019feafa, althubyani2024percy, wang2021organization, zhang2024empowering, gomes2013classifying, shrestha2024natsgd, schmidt2018introducing, ohman2020xed, santamaria2018using} integrate multimodal information such as vision, voice, and tactile signals, spanning tasks such as expression recognition, mental state detection, and fostering meaningful human-machine interactions. For \textit{\textbf{Biomedical Research}}, a broad spectrum of biological and chemical datasets~\cite{luo2022biored, li2024chemistry3d, reaxys, croft2010reactome, berman2000protein, westbrook2015chemical} serves as input for EmAI systems, accelerating biomedical analysis and discovery. Additionally, to facilitate adaptation in laboratory environments, datasets capturing instrumental and experimental scenes~\cite{zou2024benchmark, ding2021new} have been introduced to support automated experimentation.

While the existing high-quality datasets have significantly propelled EmAI research in healthcare, several notable limitations hinder their full potential. Surgical procedure datasets like Cholec80 and CaDIS are constrained by relatively insufficient sample sizes, typically comprising a few hundred videos, which restricts the ability to train robust models capable of handling the variability of surgical skills ~\cite{twinanda2016endonet, grammatikopoulou2021cadis}. Additionally, VLM datasets such as AVOS and LMOD may suffer from inconsistencies in data annotation and synchronization between visual and linguistic inputs, affecting the reliability of multi-modal policy learning~\cite{goodman2024analyzing, qin2024lmod}.
Moreover, Postoperative rehabilitation datasets such as Smart-Insole and UCI-HAPT, while rich in sensor data, often lack comprehensive contextual information necessary for developing personalized rehabilitation protocols~\cite{dai202415m, althubyani2024percy}. Additionally, biomedical research datasets such as BioRED and Reactome, while comprehensive in biological and chemical information, often face challenges related to data interoperability and standardization, impeding seamless integration with EmAI systems~\cite{luo2022biored, croft2010reactome}. Datasets that capture instrumental and experimental scenes may not fully reflect the variability of laboratory environments, thereby limiting the adaptability of automated experimental systems~\cite{zou2024benchmark, ding2021new}. Additionally, the perspective of healthcare world models is highly anticipated, but healthcare data is clearly more limited compared to that in general domains \new{for building} a comprehensive world model.

\section{Challenges and Opportunities}\label{section6}
Despite ongoing advancements, the development of Embodied AI (EmAI) systems for healthcare continues to face significant challenges. These challenges span ethical, legal, technical, and social domains, requiring comprehensive strategies to address issues such as liability, data security, system interoperability, and equitable resource allocation. Ensuring the effective integration of EmAI into healthcare also demands robust ethical oversight, clear regulatory frameworks, and scalable technical solutions, while balancing the economic and social impacts of these innovations. By overcoming these hurdles, EmAI systems can unlock their full potential to transform healthcare delivery and improve patient outcomes globally.

\subsection{Ethical and Legal Challenges}
\subsubsection{Liability and Accountability}
One of the main legal challenges with the implementation of EmAI in healthcare is determining liability when malfunctions or errors cause harm to patients. The question of who is responsible—the device manufacturer, the software developer, or the healthcare provider—becomes more complex with AI's autonomous capabilities~\cite{Cestonaro2023Defining, Terranova2024AI}. \new{It is essential to establish clear guidelines for liability and accountability.} This \new{requires} creating laws and regulations that account for the unique nature of AI, possibly introducing new legal categories to address the complexities of AI-driven outcomes~\cite{Solaiman2024Regulating, Cestonaro2023Defining}. Additionally, insurance policies and liability clauses \new{must be reexamined and potentially redesigned} to reflect AI's role in healthcare, ensuring patient protection and providing a clear path for accountability when things go wrong.
\subsubsection{Ethical Oversight}
The deployment of EmAI also raises \new{important} ethical questions, particularly around patient consent and the transparency of AI operations~\cite{price2019privacy, saheb2021mapping, luxton2022intelligent}. Ethical oversight committees are crucial for monitoring and guiding the \new{responsible use} of AI technologies. These committees should include not only ethicists and legal experts, but also technologists, healthcare providers, and patient representatives, to ensure a well-rounded evaluation of AI technologies from various perspectives. Key considerations include making sure that patients \new{fully} understand the role and function of AI in their care when giving consent, and \new{being} transparent about how AI uses patient data and influences decisions~\cite{Wang2024Ethical, Morley2020The}. Ethical oversight must also ensure that these technologies do not \new{worsen} existing healthcare disparities or introduce new forms of bias or discrimination.

\subsection{Security}
\subsubsection{Intervention Security Risks}
Beyond the extensively discussed topics of AI security~\cite{sarker2021ai, hu2021artificial, bertino2021ai}, EmAI in healthcare presents a unique challenge -- the intervention security risk. The integration of physical devices, such as surgical tools controlled by EmAI-driven decisions, into real-world healthcare scenarios introduces risks of patient harm and medical malpractice. Cyber-attacks~\cite{farooq2023cyber, giaretta2024cybersecurity, zeng2022ai}, adversarial attacks on AI algorithms~\cite{finlayson2019adversarial, qiu2019review, liang2022adversarial}, and data-poisoning attacks~\cite{alber2025medical} can lead to unpredictable outcomes, \new{which could result in physical harm to patients and threaten their safety.} Robust protection against unauthorized access and malicious interventions is therefore critical. Moreover, even without external threats, AI algorithms, knowledge databases, and robotic devices within EmAI systems must work in harmony to ensure precise and accurate actions, minimizing errors and safeguarding patient outcomes. This necessitates \new{thorough} validation, real-time monitoring, anomaly detection, and fail-safe mechanisms. While LLM agents acting as evaluators can, to some extent, enhance precision, their performance remains inadequate \new{for full reliability}~\cite{chan2023chateval, zhang2023wider, panickssery2024llm}. Lastly, comprehensive research on the scope of EmAI system application is essential to define security boundaries and ensure appropriate deployment in healthcare settings~\cite{pee2019artificial}, preventing unintended consequences and ensuring consistent performance—a relatively underexplored area. These challenges underscore the growing need for research in AI security and the precision control of medical devices.
\subsubsection{Data Protection}
Another primary concern in healthcare EmAI is ensuring the integrity and confidentiality of data, especially as EmAI systems may collect extensive patient and medical environment data. \new{Protecting this data through encryption is essential.} Advanced methods, such as homomorphic encryption, allow AI systems to process encrypted data without needing to decrypt it, thus maintaining privacy even during analysis~\cite{munjal2023systematic}. Additionally, blockchain technology can be used to create secure, unchangeable records of medical data transactions, \new{further enhancing} security across distributed networks~\cite{yaqoob2022blockchain}. Model pruning techniques~\cite{pasandi2020modeling, cheng2024survey}, which reduce model size, also hold potential for enabling offline EmAI systems. These technologies help protect against unauthorized access and data breaches, which are critical in patient data management.
\subsubsection{Compliance Systems}
Compliance with legal and ethical standards is crucial when deploying AI in healthcare. Developing \new{AI-powered} compliance systems can help healthcare providers navigate complex regulatory frameworks like the GDPR in Europe or HIPAA in the United States~\cite{voigt2017eu, ness2007influence}. These systems must be flexible, \new{enabling them to adapt to new regulations and policies as they emerge}. \new{By leveraging AI, compliance systems can automatically update and audit data usage across platforms, ensuring ongoing adherence to data protection laws}~\cite{Arbabi2023A}. This not only supports responsible data management but also fosters trust with patients and regulatory bodies.

\subsection{Technical Challenges}
\subsubsection{Accuracy and Explainability Enhancement}
AI models, particularly those involved in diagnostics and patient monitoring, must perform with high precision to be trusted by healthcare providers and patients alike. Techniques such as deep learning have shown potential in improving the accuracy of AI systems, but they require \new{vast amounts} of high-quality, diverse data~\cite{liang2019evaluation}. Enhancing the reliability of these systems also involves implementing robust validation frameworks that simulate a wide range of clinical scenarios to test AI responses and avoid LLM hallucinations~\cite{Hadjiiski2022AAPM, Elul2021Meeting}. Continuous learning algorithms can be utilized to update and refine AI models \new{as new data and outcomes become available}, gradually improving their accuracy and reliability over time. 

Explainability in EmAI is particularly critical in healthcare, as it clarifies how these systems perceive, make decisions, and take action—boosting both transparency and trust. For example, a surgical EmAI system should clearly present patient condition data and explain each surgical step, such as adjusting a scalpel angle due to abnormal blood flow detection. \new{These detailed explanations are essential for building trust among clinicians, especially during the early stages of EmAI adoption when acceptance is still growing.} To enhance explainability, methods such as multimodal attribution analysis~\cite{jain2023maea, sun2024review}, feature visualization~\cite{chefer2021transformer}, multimodal knowledge graphs~\cite{song2024scene, zhu2022multi}, and causal inference~\cite{gat2023faithful, tan2023causal, ji2023benchmarking} can be employed. \new{Additionally, leveraging the conversational capabilities of LLMs can provide grounded, theory-based explanations, revealing causal relationships in medical decision-making and helping clinicians navigate complex medical scenarios.}

\subsubsection{Interoperability of Devices}
Ensuring that different AI-powered devices and existing healthcare systems work together smoothly is a major challenge. For EmAI to be effective, it needs to integrate and communicate seamlessly with various healthcare IT systems and medical devices~\cite{Chatterjee2022HL7, Duda2022HL7}. \new{Standardizing} data formats and communication protocols is key to making this possible. Initiatives like Health Level Seven International (HL7) and Fast Healthcare Interoperability Resources (FHIR) standards are \new{crucial} in driving this integration~\cite{dolin2006hl7, bender2013hl7}. By enabling AI systems to interact with each other and with \new{existing systems}, healthcare providers can offer more coordinated and efficient care. Additionally, \new{developing APIs and open standards} can simplify data exchange and functionality across different platforms and devices, making AI solutions more scalable and useful in healthcare settings. \new{It’s also important to create systems that work with different types of robots and can combine various capabilities—such as segmenting CT scans, X-rays, or engaging in dialogue—into a single system. This suggests the need for AI with modular designs that can integrate multiple functions and handle diverse data efficiently.}

\subsubsection{Human Digital twin and World Model for Simulation}
The integration of human digital twin and world models, particularly video generation models and 3D generation models, into the biomedical domain holds immense potential~\cite{cao2024medical,sun2024bora,khader2023denoising}. These models can enhance the understanding of diagnostic and surgical decisions and improve their generalizability from virtual to physical environments. Recent advances in medical video generation have expanded to various applications, including simulating disease progression in X-rays, fundus images, and skin images~\cite{cao2024medical}, generating surgical videos~\cite{cho2024surgen,chen2024surgsora}, and creating text-to-video simulation for diverse medical imaging modalities~\cite{sun2024bora,li2024endora}. The application of these medical video generation models has the potential to improve how we build EmAI in healthcare, though it is important to be aware of its current limitations. First, the availability of trainable medical video datasets is limited, which constrains the diversity and robustness of these models~\cite{cao2024medical,kang2024farvideogenerationworld}. 
Second, the simulated videos generated by current models may lack reliability in corner cases, and generating long videos remains computationally intensive~\cite{chen2024surgsora}. 
Moreover, the inclusion of multi-view or 3D spatial-temporal information~\cite{zhang2025physdreamer,li2024next,chen2024revolution} is essential for applications like surgical operations, yet most existing models are limited to 2D video generation. In the future, it would be important to build customized 3D simulators and world models in various medical application scenarios~\cite{zhu2024sora,wu2024robomind,agarwal2025cosmos}.

\subsubsection{Stronger Generalization Capabilities}
The future of EmAI in healthcare lies in its potential to seamlessly integrate physical and AI capabilities, transforming the way healthcare systems interact with patients and their environments. Currently, most research in this field has focused on discrete applications—such as robotic-assisted surgery or rehabilitation—yet the broader frontier is in developing EmAI for more generalized settings.  
To this end, the exploration of spatial intelligence, multi-modality foundational models, large language models, world models, and physical AI is crucial~\cite{xiang2024language}. Additionally, evolutionary algorithms that autonomously tune hyperparameters and support continuous learning also help models evolve and better adapt post-deployment in diverse environments. \new{By enabling a network of autonomous agents to coordinate and share real-time insights, EmAI systems can further develop more adaptive and scalable generalization capabilities through the power of collective intelligence.}

\subsection{Human-Machine Interaction}
\subsubsection{User Training and Education}
The introduction of EmAI into healthcare settings \new{requires} significant user training and education to ensure that healthcare providers can effectively interact with and leverage these technologies~\cite{labkoff2024toward}. Training programs must be developed to cater to a diverse range of skills and familiarity with technology among healthcare workers. These programs should focus on the operational aspects of the technology, as well as on understanding the AI’s decision-making process, to build trust and confidence in AI systems~\cite{Smith2023Clinicians}. Furthermore, continuous education needs to be provided to keep pace with the rapid evolution of AI technologies, ensuring that clinicians remain competent in \new{using} these advanced tools.
\subsubsection{Workflow Integration}
\new{Integrating EmAI systems into healthcare workflows is challenging but necessary.} These systems must be designed to complement and enhance existing healthcare practices without causing disruption~\cite{Catchpole2022Human, Andras2019Artificial}. For instance, EmAI tools such as robotic assistants, smart prosthetics, and augmented reality systems for surgery or patient care need to be seamlessly integrated into hospital routines. This integration requires careful planning and adjustment of clinical pathways, involving both the redesign of physical spaces and the modification of procedural protocols to accommodate the new technologies. The goal is to ensure that these AI systems improve efficiency and patient outcomes without adding unnecessary complexity to the clinicians’ workload.

\subsection{Economic and Social Impact}
\subsubsection{Cost-Benefit Analysis}
\new{Deploying} EmAI technologies in underdeveloped areas \new{comes with} significant costs, including the initial investment in technology, ongoing maintenance, and training of clinicians to use these systems effectively~\cite{aderibigbe2023artificial}. However, the benefits can be substantial, \new{as EmAI can revolutionize healthcare delivery by offering advanced diagnostic tools, telemedicine, and robotic assistance in areas with a shortage of clinicians.} A detailed cost-benefit analysis must consider the long-term savings in healthcare costs due to improved disease prevention, diagnostic accuracy, and treatment outcomes. This analysis should also evaluate how these technologies can extend healthcare reach, reduce the burden on understaffed clinics, and improve patient care in remote areas. The ultimate goal is to determine whether the high initial costs are justified by the anticipated improvements in healthcare quality and accessibility.

\subsubsection{Social Resource Allocation}
The introduction of EmAI in underdeveloped regions also poses questions about the allocation of social resources. Decisions need to be made about how to distribute limited resources—whether to invest in advanced AI technologies or to allocate funds to more basic healthcare needs like vaccinations and clean water. This becomes a complex issue of prioritization, where the potential long-term benefits of AI must be weighed against immediate healthcare necessities. It is crucial to engage community stakeholders in these decisions to ensure that the allocation of resources aligns with the actual needs and priorities of the community. Moreover, policymakers must consider how these investments in AI technology might affect social equity, \new{ensuring these technologies reduce health disparities rather than widening them}~\cite{holzmeyer2021beyond}.

\section{Conclusion}
Embodied AI (EmAI) for healthcare represents a transformative paradigm, integrating artificial intelligence with physical systems to deliver personalized, scalable, and adaptive solutions across diverse medical domains. By combining capabilities in perception, action, decision-making, and memory, EmAI systems have emerged as robotic clinical assistants, companion caregivers, autonomous diagnostic tools, and biomedical researchers, demonstrating significant potential to improve patient outcomes, reduce the burden on healthcare providers, and enhance access to medical services. We have also explored the intelligent levels of EmAI systems, highlighting that their development remains in its early stages while providing guidance for future advancements. However, the field still faces substantial challenges, including concerns around data privacy, system reliability, ethical considerations, limited application scope, and the complexities of integration into existing clinical workflows.
Future research should focus on addressing these challenges while advancing multimodal sensing, human-AI interaction, and adaptive learning capabilities to ensure safe and effective deployment in real-world settings. Despite its infancy, EmAI holds great promise to transform healthcare by overcoming technical and ethical barriers, ultimately enhancing patient care and advancing healthcare systems.

\section{Acknowledgement}
This research was partially supported by the High Performance Computing Center of Central South University and the Fundamental Research Funds for the Central Universities of Central South University. Jintai Chen's contribution was supported by internal funding from the Hong Kong University of Science and Technology (Guangzhou).
{
\bibliographystyle{elsarticle-num}
\bibliography{main}
}

\end{document}